\documentclass{article}

\PassOptionsToPackage{numbers, compress}{natbib}

\usepackage[main, final]{neurips_2026}

\usepackage[utf8]{inputenc} 
\usepackage[T1]{fontenc}    
\usepackage{CJKutf8}        
\usepackage{hyperref}       
\usepackage{url}            
\usepackage{graphicx}       
\usepackage{amsmath}        
\usepackage{booktabs}       
\usepackage{amsfonts}       
\usepackage{nicefrac}       
\usepackage{microtype}      
\usepackage{xcolor}         
\usepackage{pifont}

\definecolor{bestbg}{HTML}{CFE8FF}     
\definecolor{secondbg}{HTML}{FFE8CC}   
\usepackage{xcolor,colortbl,booktabs,multirow}
\usepackage{multirow}          
\usepackage{pifont}            
\newcommand{\cmark}{\ding{51}} 
\newcommand{\xmark}{\ding{55}} 

\title{Why Struggle with Continuous Latents? Interpretable Discrete Latent Reasoning via Rendered Compression}

\newcommand{\equal}{\textsuperscript{*}} \newcommand{\corrauth}{\textsuperscript{\textdagger}} \newcommand{\sjtu}{\textsuperscript{1}} \newcommand{\tongji}{\textsuperscript{2}}

\author{ \textbf{Shuochen Chang}\sjtu\equal \quad \textbf{Qingyang Liu}\sjtu\equal \quad \textbf{Shaobo Wang}\sjtu \quad \textbf{Bingjie Gao}\sjtu \quad \textbf{Qianli Ma}\sjtu \\ \textbf{Haonan Zhao}\sjtu \quad \textbf{Yibo Miao}\sjtu \quad \textbf{Yulin Sun}\sjtu \quad \textbf{Zelin Peng}\sjtu \quad \textbf{Jiangtong Li}\tongji\corrauth \quad \textbf{Li Niu}\sjtu\corrauth \\ \sjtu Shanghai Jiao Tong University \quad \tongji Tongji University \\ \equal Equal contribution. \quad \corrauth Corresponding authors. \\ \texttt{\{csc1332741686, narumimaria, ustcnewly\}@sjtu.edu.cn} \\ \texttt{jiangtongli@tongji.edu.cn} }

\begin{document}

\begin{CJK*}{UTF8}{gbsn}

\maketitle

\begin{abstract}
Large language models achieve high reasoning performance via explicit chain-of-thought and reinforcement learning, but require long output sequences and extended inference time.
Latent reasoning reduces this cost by shifting computation into a latent space; however, continuous latent methods are hard to train, suffering from unstable and uninterpretable reasoning trajectories.
We argue these issues stem from a misalignment between continuous-space reasoning and discrete symbolic supervision, as continuous states lack explicit anchors for step-by-step alignment.
To resolve this, we propose \textbf{Discrete Latent Reasoning~(DLR)}, the first method that converts continuous latent states into explicit discrete tokens.
Inspired by render-based compression, we render textual chains of thought into images, extract visual features, and construct a discrete latent vocabulary via clustering-based fine-tuning.
Expanding the vocabulary and output head enables standard autoregressive modeling over both natural language and latent tokens, supporting pretraining alignment, SFT, and RL.
Experiments on five reasoning benchmarks and two model series~(Qwen3-VL and LLaMA-3) confirm that \textbf{DLR} outperforms prior latent reasoning baselines with up to \textbf{20$\times$ compression}.
Furthermore, the learned latent trajectories retain an interpretable semantic structure. 
Overall, discrete latent tokens provide a controllable and interpretable basis for efficient latent reasoning.
Code is available at \url{github.com/Miraclecsc/Discrete-Latent-Reasoning}.
\end{abstract}

\section{Introduction}
Large language models~(LLMs) have achieved high reasoning performance via chain-of-thought~(CoT) and reinforcement learning~(RL)~\cite{wei2022chain,hsieh2023distilling,deepseekai2025deepseekr1,stepfun2026inside}.
However, they often require large token budgets and extended inference time to solve difficult problems.
Latent reasoning offers an efficient alternative by shifting intermediate computation into a high-bandwidth latent space, preserving uncertainty structures while reducing verbalization costs.
This approach is supported by recent visual intermediate representations~\cite{wei2026deepseekocr2,wang2026rot,chen2026stdec,li2026sage,wang2026solving}, which indicate that reasoning can be effectively compressed in image space.

Recent analyses of representative latent systems~\cite{li2026imagination,zhang2026nitp,zhang2026xhc} reveal that continuous latent tokens are highly susceptible to representation collapse, often remaining overly similar and yielding limited impact on the final answer~\cite{yang2025mirage,li2025lvr,wang2025monet,qin2026sstoken}. 
These issues expose the core vulnerabilities of continuous latent models, which primarily suffer from three critical limitations.
First, these models introduce a major \textbf{training bottleneck}.
During supervised fine-tuning~(SFT), continuous latent states lack explicit discrete anchors analogous to text tokens, making teacher forcing difficult and frequently requiring sequential unrolling of latent trajectories, which reduces efficiency and compatibility with existing SFT and RL methods~\cite{hao2024coconut,wu2025parallel}.
During RL, the absence of explicit step boundaries complicates the assignment of process-level rewards and disrupts stable credit assignment over latent transitions~\cite{lightman2024lets,williams2026prioritize}.
Second, they frequently display \textbf{unstable reasoning dynamics}.
Since the latent channel is continuous and only weakly constrained, local errors can accumulate across steps, causing instability, representation collapse, or deviation from decision-relevant evidence~\cite{zou2026capabilities,wei2025simcot}.
Moreover, unlike explicit textual reasoning, latent states lack strong language modeling priors from natural language tokens, limiting their stability and compositionality~\cite{deng2025latent,cui2026weakstrong}.
Third, latent reasoning is largely opaque and suffers from \textbf{poor interpretability}, making the reasoning process hard to inspect, verify, or correct in reliability-critical settings~\cite{chang2026unlockingblackbox,korbak2025cotmonitorability,qin2026multi,}.

\begin{figure}[t]
  \centering
  \includegraphics[width=0.96\linewidth]{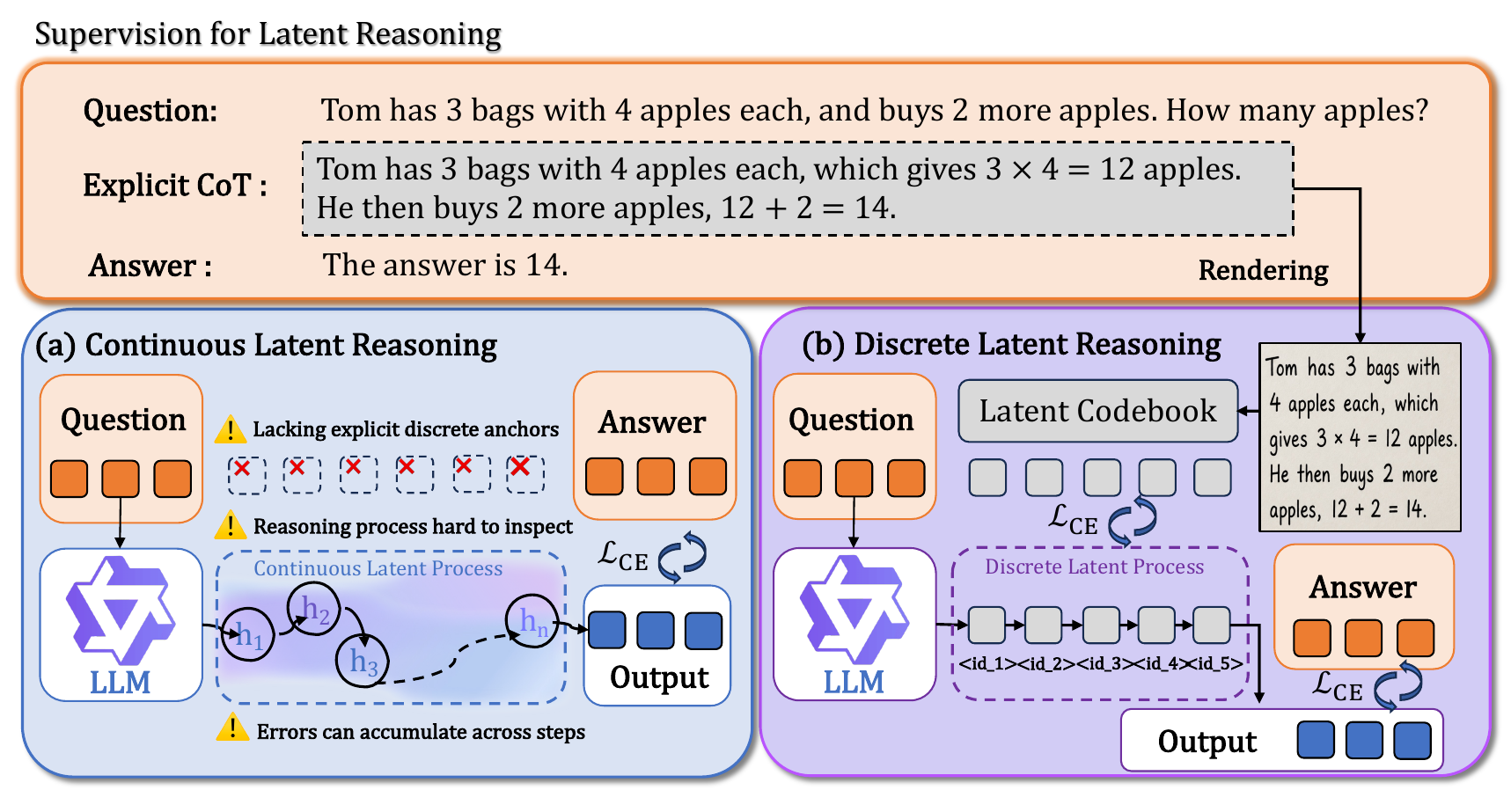}
  \caption{Continuous latent reasoning suffers from a mismatch between its intermediate states and discrete symbolic supervision, resulting in a training bottleneck, unstable reasoning dynamics, and poor interpretability.
  Our approach resolves this by introducing discrete latent tokens as anchors within the reasoning process, enabling standard autoregressive training and controllable latent reasoning.}
  \label{fig:intro-discrete-latent}
\end{figure}

The above limitations stem from a \textbf{misalignment between continuous-space reasoning and discrete symbolic supervision}.
Unlike standard autoregressive methods built upon discrete vocabularies~\cite{brown2020language,ouyang2022training}, continuous latent reasoning lacks explicit anchors, making intermediate states difficult to supervise, inspect, or correct~\cite{deng2025latent,cui2026weakstrong}.
To retain computational efficiency while enabling stable and interpretable training, we propose resolving this by \textbf{discretizing latent states}, \emph{i.e.}, transforming implicit reasoning steps into explicit latent tokens directly compatible with standard language modeling.

In this work, we propose \textbf{Discrete Latent Reasoning~(DLR)}~(Fig.~\ref{fig:intro-discrete-latent}), a novel method that converts continuous latent states into discrete latent tokens for latent reasoning.
Inspired by the render-compression strategy of DeepSeek-OCR~\cite{wei2026deepseekocr2}, we design a \textbf{stochastic latent codebook construction module}, which renders textual chains of thought into images and leverages optical compression to extract dense, semantically structured visual features for latent tokenization.
We then introduce a stochastic clustering-based fine-tuning mechanism to construct a discrete latent vocabulary, where each token corresponds to an interpretable semantic latent unit.
To enable autoregressive modeling over these units, we propose an \textbf{augmented latent language model training pipeline}, which expands both the vocabulary and the output head of LLMs, allowing them to process and generate sequences in an augmented space comprising both natural language and latent tokens.
The training proceeds in three stages: \textit{\textbf{(i) latent-text alignment}} to synchronize the latent embeddings in the LLM space; \textit{\textbf{(ii) latent SFT}} to teach the model to generate correct latent trajectories; and \textit{\textbf{(iii) latent RL}} to optimize the latent-token policy via outcome and process rewards.
This allows latent reasoning to be optimized via next-token prediction, supports process-level rewards over discrete steps, reduces error accumulation through structured anchors, and improves interpretability via an auxiliary decoder.



To validate the effectiveness of our approach, we fine-tune both large language models~(LLaMA3) and large multimodal models~(Qwen3-VL) on the GSM8K-Aug-train~\cite{cobbe2021training} split.
Compared to previous latent reasoning baselines such as RoT~\cite{wang2026rot}, ReGuLaR~\cite{wang2026regular}, and CODI~\cite{shen2025codi}, our method achieves consistent accuracy improvements across GSM8K~\cite{cobbe2021training}, GSM-Hard~\cite{gao2023pal}, SVAMP~\cite{patel2021nlp}, MultiArith~\cite{roy2015solving}, and MATH-500~\cite{hendrycks2021math}.
Moreover, we conduct experiments on the scalability of \textbf{DLR} and the interpretability of the latent trajectories, finding that the model can scale from 1B to 8B LLMs while retaining a meaningful degree of interpretability.
Our contributions are summarized as:



\begin{itemize}

    \item We propose a \textbf{Stochastic Latent Codebook Construction} pipeline that discretizes continuous reasoning states into semantically recoverable latent tokens through rendered CoT, providing interpretable latent representations with up to \textbf{20$\times$ compression}.

    \item We design \textbf{an Augmented Latent LM Training Pipeline}, which enables standard next-token prediction during SFT and supports process-reward RL. This design maintains semantic fidelity while significantly improving training stability.

    \item Together, we present \textbf{Discrete Latent Reasoning~(DLR)}, the first discrete latent reasoning framework, and show its superiority over latent reasoning baselines on five reasoning benchmarks with improved trajectory interpretability and scalability.

\end{itemize}

\section{Related Works}
\subsection{Continuous Latent Reasoning}
Latent reasoning improves efficiency by replacing verbose text with dense continuous states. 
For instance, Coconut~\cite{hao2024coconut} introduces autoregressive reasoning in latent space by utilizing final hidden states.
Subsequent works advance this approach: CODI~\cite{shen2025codi} aligns trajectories via self-distillation, CoLaR~\cite{tan2025colar} uses token-guided dynamic compression, and SIM-CoT~\cite{wei2025simcot} incorporates supervised decoding signals.
Additional research investigates KV-cache distillation, superposition, and looped computation~\cite{kuzina2025kava,deng2025latent,zhu2025looped}.
Concurrently, several methods extend these techniques to multimodal tasks, preserving visual cues within latent representations~\cite{yang2025mirage,li2025lvr,wang2025monet,liu2025reasoningwithin,jeon2026valr}.
Despite their differences, these methods all treat continuous latent tokens as the medium for intermediate reasoning semantics.

\subsection{Chain-of-Thought Compression and Visual Compression}
As reasoning models produce increasingly long chains of thought, another line of work investigates how to compress these traces with minimal loss of performance.
In the text domain, existing methods often employ preference optimization~\cite{zhang2025tokensqueeze} or reward signals that encourage conciseness~\cite{wang2025efficient}.
Inspired by the optical compression of DeepSeek-OCR~\cite{wei2025deepseekocr}, recent approaches use rendered-text compression to connect explicit CoT with latent computation.
Specifically, Render-of-Thought~\cite{wang2026rot} uses visual embeddings of rendered rationales for supervision; OneLatent~\cite{lv2026onelatent} compresses these into a single token; and ReGuLaR~\cite{wang2026regular} treats rendered CoT as a visual-semantic prior.
These studies indicate that effective compression requires encoding reasoning steps into denser, and continuous representations.

\subsection{Unified Model}
Unified multimodal models represent heterogeneous modalities within a single generative architecture.
Chameleon~\cite{chameleonteam2024chameleon} models interleaved sequences via early fusion, while Emu3~\cite{wang2024emu3} extends next-token prediction across modalities using a unified token space.
Show-o and Janus~\cite{xie2024showo,chen2024janus} further advance this concept through varied visual encoding strategies.
Conceptually, these models align closely with vector-quantized latent modeling: VQ-VAE establishes how continuous representations can be discretized into codebook entries that retain semantic structure while remaining compatible with autoregressive symbolic modeling~\cite{oord2017neural}.
Building on this principle, rather than unifying external modalities, our method discretizes internal reasoning states into vocabulary-level tokens, enabling latent computation directly within standard autoregressive language modeling.

\section{Methods}
\label{sec:methods}

\subsection{Overview}
\label{sec:method_overview}

\begin{figure}[t]
  \centering
  \includegraphics[width=\linewidth]{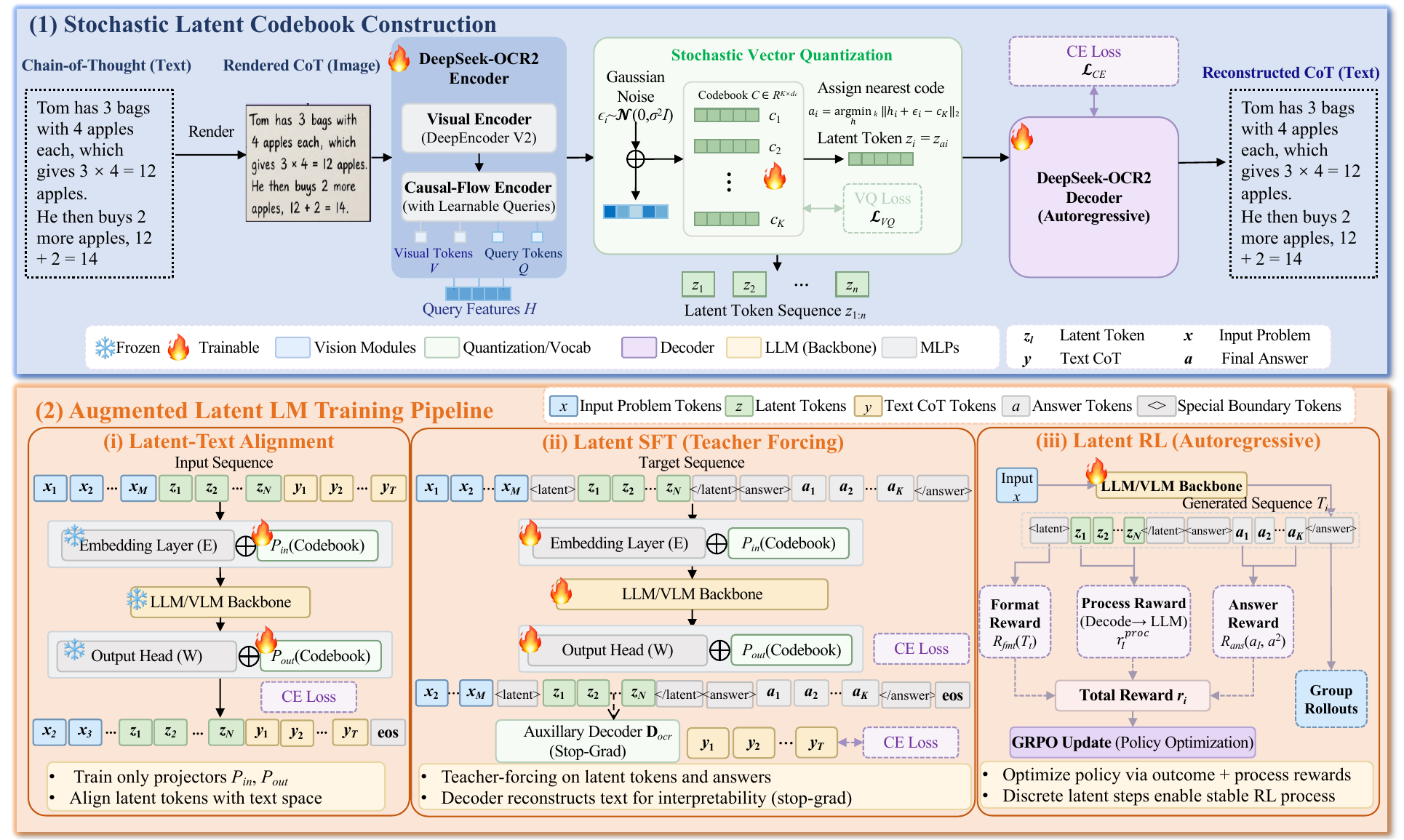}
  \caption{Overview of \textbf{DLR}. \textbf{Top}: we render chain-of-thought traces into images, encode them with a DeepSeek-OCR2-based visual compressor, and learn a stochastic vector-quantized latent codebook jointly with an OCR decoder to obtain semantically recoverable latent tokens. \textbf{Bottom}: the learned latent vocabulary is projected into the LLM token space and used to train an augmented latent language model through latent-text alignment, supervised fine-tuning, and reinforcement learning.}
  \label{fig:method}
\end{figure}

\textbf{Discrete Latent Reasoning (DLR)} bridges the optimization gap between continuous latent reasoning and standard discrete autoregressive LLM training.
Instead of relying on continuous vectors, \textbf{DLR} represents intermediate reasoning states as explicit discrete latent tokens within a unified sequence.
As illustrated in Fig.~\ref{fig:method}, our method features two main components: a \textbf{stochastic latent codebook construction} module and an \textbf{augmented latent LM training procedure}.

The first component constructs a stable latent vocabulary by extracting and discretizing features from rendered CoT traces.
We encode these visual traces using a DeepSeek-OCR2 encoder~\cite{wei2026deepseekocr2}, whose learnable queries extract causally ordered, structurally stable features ideal for vector quantization.
These dense visual features are then mapped into a vector-quantized codebook.
Joint training with the DeepSeek-OCR2 decoder ensures these discrete tokens retain high semantic fidelity and remain textually reconstructable.
This process yields a structured latent vocabulary, denoted as $\mathcal{V}_{\mathrm{lat}} = \{ z_1, \ldots, z_K \}$, ready for LLM integration.


The second component trains the LLM to autonomously generate and reason over these latent tokens alongside natural language.
We expand the text vocabulary $\mathcal{V}_{\mathrm{text}}$ by using lightweight projectors to map codebook vectors into the LLM's dimensionality, yielding latent embedding and output matrices $E_{\mathrm{lat}}$ and $W_{\mathrm{lat}}$.
By merging the vocabularies ($\mathcal{V} = \mathcal{V}_{\mathrm{text}} \cup \mathcal{V}_{\mathrm{lat}}$) and concatenating their weights ($E = [E_{\mathrm{text}}; E_{\mathrm{lat}}]$, $W = [W_{\mathrm{text}}, W_{\mathrm{lat}}]$), the model naturally performs standard next-token prediction $p(s_t \mid s_{<t}) = \mathrm{softmax}(W^\top u_t)$ over the unified space.
Training proceeds in three stages: \textit{\textbf{(i) latent-text alignment}} to synchronize embeddings; \textit{\textbf{(ii) latent SFT}} to teach latent trajectory generation; and \textit{\textbf{(iii) latent RL}} to optimize the policy via process and outcome rewards.

\subsection{Stochastic Latent Codebook Construction}\label{sec:latent_codebook}


The latent codebook converts rendered reasoning traces into reusable discrete tokens while preserving semantic recoverability throught feature extraction, vector quantization, and codebook optimization.

\textbf{Causal-Flow Rendered Feature Extraction.}
Given a textual chain-of-thought $y = (y_1,\ldots,y_T)$, we first render it into an image $I = \mathcal{R}(y)$ and encode it using a DeepSeek-OCR2 encoder~($\mathbf{E}_{\mathrm{vis}}$). 
The encoder maps $I$ into compressed visual tokens $V = \mathbf{E}_{\mathrm{vis}}(I) \in \mathbb{R}^{N_v \times d_c}$. 
To extract causally ordered semantics, we maintain a pool of position-aware learnable queries and dynamically truncate it to match the visual token length, yielding $Q \in \mathbb{R}^{N_v \times d_c}$.
The concatenated sequence $[V;Q]$ is processed by a causal transformer $\mathbf{E}_{\mathrm{cf}}$, and we select the query-side outputs as our continuous reasoning features $H = [h_1,\ldots,h_{N_v}] = \Pi_Q(\mathbf{E}_{\mathrm{cf}}([V;Q]))$, where $\Pi_Q$ retrieves the last $N_v$ positions.
This yields a causally ordered, visually compressed representation of the trace~\cite{wei2026deepseekocr2}.

\textbf{Stochastic Vector Quantization.} 
We introduce a vector-quantized codebook $C = \{c_1,\ldots,c_K\}$ with $c_k \in \mathbb{R}^{d_c}$, which is warm-started via clustering on reasoning corpora to prevent early collapse.
Each code vector $c_k$ corresponds to a discrete latent token $z_k$ in the vocabulary.
To map continuous features $h_i$ to discrete latent tokens with stability while improving codebook utilization, we employ stochastic quantization by injecting Gaussian perturbation $\epsilon_i \sim \mathcal{N}(0,\sigma^2 \mathbf{I})$ into $h_i$.
The discrete index $a_i$ is then assigned via minimum $L_2$ distance:
\begin{equation}
    a_i = \arg\min_{k \in \{1,\ldots,K\}} \| h_i + \epsilon_i - c_k \|_2.
\end{equation}
The continuous feature is replaced by the corresponding code vector to form the quantized feature $\hat{h}_i = c_{a_i}$, and is then mapped to the discrete latent token $z_i = z_{a_i}$.
This process yields the quantized embedding sequence $\hat{H} = [\hat{h}_1,\ldots,\hat{h}_{N_v}]$ and the discrete token sequence $z_{1:N_v}$.

\textbf{Curriculum-based Codebook Optimization.} 
The codebook is trained to ensure that the discrete tokens retain interpretable semantics.
To allow gradients to propagate through the discrete assignment, we use the straight-through estimator to form the differentiable quantized embeddings $\hat{h}_i^\mathrm{st} = h_i + \mathrm{sg}[\hat{h}_i - h_i]$, with the sequence denoted as $\hat{H}^\mathrm{st}$, where $\mathrm{sg}[\cdot]$ is the stop-gradient operator.
The decoder of DeepSeek-OCR2~($\mathbf{D}_\mathrm{ocr}$) is trained to reconstruct the original text $y$ from $\hat{H}^\mathrm{st}$ using
\begin{equation}
  \mathcal{L}_\mathrm{CE}(\hat{H}^\mathrm{st}, y) = - \sum_{t=1}^{T} \log \mathbf{D}_\mathrm{ocr}(y_t \mid y_{<t}, \hat{H}^\mathrm{st}).
\end{equation}
The discrete codebook assignments are learned via a standard vector-quantization loss:
\begin{equation}
    \mathcal{L}_\mathrm{VQ} = \sum_{i=1}^{N_v} \|\mathrm{sg}[h_i] - \hat{h}_i\|_2^2 + \beta \sum_{i=1}^{N_v} \|h_i - \mathrm{sg}[\hat{h}_i]\|_2^2.
\end{equation}
To stabilize early training, we employ a two-branch curriculum where the decoder receives both continuous features $H$ and quantized features $\hat{H}^\mathrm{st}$.
The step-$e$ training loss is:
\begin{equation}
    \mathcal{L}_\mathrm{cb}^{(e)} = \mathcal{L}_\mathrm{CE}(\hat{H}^\mathrm{st}, y) + \alpha(e)\mathcal{L}_\mathrm{CE}(H, y) + \lambda_\mathrm{VQ} \mathcal{L}_\mathrm{VQ}.
\end{equation}
The annealing factor $\alpha(e)$ decays from 1 to 0 over the first epoch. 
After convergence, the continuous branch is entirely discarded, forcing the OCR decoder to act purely as an interpretability module reading from the discrete latent trajectory.

\subsection{Augmented Latent Language Model Training Pipeline}
\label{sec:training_framework}

After constructing the latent codebook, the LM is trained over the augmented vocabulary $\mathcal{V}_\mathrm{text} \cup \mathcal{V}_\mathrm{lat}$, transitioning through three stages: latent-text alignment, latent SFT, and latent RL.

\subsubsection{Latent-Text Alignment}
\label{sec:latent_text_alignment}

The goal of this stage is to align the newly introduced latent tokens with the pretrained text manifold.
We freeze the LLM backbone and update only the lightweight input/output projectors. 
Given a reasoning trace $y$, we quantize its rendered features into a latent sequence $z_{1:n}$. 
We then construct a concatenated sequence $s = [z_{1:n}, y_{1:T}]$ and optimize the projectors using a standard autoregressive cross-entropy loss over the entire sequence. 
This enforces the latent prefix to encode sufficient semantic context, enabling the frozen LLM to continue generating the corresponding explicit text.

\subsubsection{Latent Supervised Fine-Tuning}
\label{sec:latent_sft}

During SFT, the model learns to autonomously generate latent reasoning trajectories conditioned on task inputs. 
To support structured generation and format verification during the subsequent RL stage, we introduce special boundary tokens.
For a problem $x$, reasoning trace $y$, and answer $a$, we construct the target sequence using the trained codebook as:
\begin{equation}
    s = [x; \mathtt{<latent>}; z_{1:n}; \mathtt{</latent>}; \mathtt{<answer>}; a_{1:M}; \mathtt{</answer>}].
\end{equation}
The LLM is optimized via standard next-token cross-entropy over the entire generated portion, encompassing latent tokens, final answer, and boundaries:
\begin{equation}
    \mathcal{L}_\mathrm{SFT} = - \sum_{t=|x|+1}^{|s|} \log p_\theta(s_t \mid s_{<t}).
\end{equation}
To preserve the interpretability of latent trajectories without compromising training stability, we introduce a dual-branch parallel training.
While the LLM undergoes teacher-forced training on $z_{1:n}$, the continuous input embeddings of the latent tokens, denoted as $G(z_{1:n}) = [E_\mathrm{lat}(z_1), \ldots, E_\mathrm{lat}(z_n)]$, are also fed into the auxiliary latent decoder $\mathbf{D}_\mathrm{ocr}$, which is trained to reconstruct the text $y$:
\begin{equation}
    \mathcal{L}_\mathrm{dec} = - \sum_{t=1}^{T} \log p_{\mathbf{D}_\mathrm{ocr}}\big(y_t \mid y_{<t}, \mathrm{sg}[G(z_{1:n})]\big).
\end{equation}
By applying a stop-gradient operator ($\mathrm{sg}[\cdot]$) at the LLM interface, we structurally isolate the two processes. 
This ensures the LLM independently masters latent reasoning efficiency, while the auxiliary decoder translates these latent states back into interpretable steps without interfering with the LLM's representation space.

\subsubsection{Latent Reinforcement Learning}
\label{sec:latent_rl}

In the final stage, we optimize the latent policy $\pi_\theta$ to refine accuracy and intermediate structural coherence. Unlike SFT, the full response trajectory $\tau_i \sim \pi_\theta(\cdot \mid x)$ is generated freely without teacher forcing. 
We parse $\tau_i$ based on the boundary tokens to extract the latent trajectory $z^{(i)}_{1:n_i}$ and the predicted answer $a^{(i)}_{1:M_i}$. We evaluate each rollout using a composite reward function.

To ensure the quality of the latent trajectories, we propose the \textbf{Process Alignment Reward}. 
Instead of only rewarding the final answer, we also leverage the auxiliary decoder to translate the latent trajectory into an interpretable trace $\hat{y}^{(i)} = \mathbf{D}_\mathrm{ocr}\big(G(z^{(i)}_{1:n_i})\big)$.
We then prompt a frozen evaluator LLM ($\pi_\mathrm{eval}$, \emph{e.g.}, the initial instruction-tuned checkpoint before latent-text alignment) with this decoded trace to predict a verification answer $\tilde{a}^{(i)}$, which is compared against the ground-truth $a^*$:
\begin{equation}
    \tilde{a}^{(i)} \sim \pi_\mathrm{eval}(\cdot \mid x, \hat{y}^{(i)}), \quad
    r_i^\mathrm{proc} = R_\mathrm{ans}(\tilde{a}^{(i)}, a^*).
\end{equation}
This mechanism provides a dynamic, semantic-level verification of the latent reasoning process. 
The total reward for rollout $i$ combines this process reward with the final answer correctness $R_\mathrm{ans}$ and a format verification score $R_\mathrm{fmt}$. 
Specifically, $R_\mathrm{fmt}$ ensures the model strictly adheres to the structural constraints enforced during SFT by verifying the exact presence and ordering of the boundary tokens:
\begin{equation}
    r_i = R_\mathrm{ans}(a^{(i)}, a^*) + \lambda_\mathrm{proc} r_i^\mathrm{proc} + \lambda_\mathrm{fmt} R_\mathrm{fmt}(\tau_i).
\end{equation}

With the composite reward defined, we optimize the policy using the Group Relative Policy Optimization~(GRPO) algorithm~\cite{deepseekai2025deepseekr1}. Following standard GRPO practices, we apply group-based advantage normalization, clipped probability ratios to prevent destructive policy updates, and a KL-divergence penalty against the reference policy $\pi_\mathrm{ref}$. This concise setup maximizes the expected reward while maintaining training stability, avoiding the need for a separate value model.

\section{Experiments}

\subsection{Experimental Setup}
\label{sec:experimental_setup}

\textbf{Training Data.}
DLR has two separate data pipelines: one for latent codebook training on rendered CoT images, and another for latent LLM training on mixed text-latent sequences~(details in App.~\ref{sec:appendix_data:train}).\\
For \textbf{latent codebook training}, we aggregate $\sim\!1$M reasoning traces from GSM8K-Aug~\cite{deng2023implicit}, MATH~\cite{hendrycks2021math}, and MathX-5M~\cite{modotte2026mathx5m}. 
Textual CoTs~(up to $2,048$ tokens) are rendered into adaptive square RGB images~(up to $1024 \times 1024$ resolution), producing a maximum of $256$ visual tokens.
This achieves an $\sim\!8-20\times$ compression ratio, efficiently shifting the computational reasoning overhead into the dense latent space.
For \textbf{latent LM training}, we construct two configurations. 
The \emph{base} setup utilizes $385$K GSM8K-Aug~\cite{deng2023implicit} samples, allocating all $385$K for latent-text alignment, $300$K for SFT, and the remaining $85$K for RL.
To evaluate generalization, a \emph{scaled} setup distributes $\sim\!12$K MATH~\cite{hendrycks2021math} samples across the same three training stages.


\textbf{Evaluation Data.}
For the \emph{base} configuration, we evaluate zero-shot performance on four standard arithmetic benchmarks: GSM8K~($1{,}319$)~\cite{cobbe2021training}, GSM-Hard~($1{,}319$)~\cite{gao2023pal}, SVAMP~($1{,}000$)~\cite{patel2021nlp}, and MultiArith~($600$)~\cite{roy2015solving}. 
The \emph{scaled} configuration is evaluated on the harder MATH-500 benchmark~($500$)~\cite{hendrycks2021math}. 
All evaluations are strictly designed to omit explicit chains-of-thought in the zero-shot setting~(further details are provided in App.~\ref{sec:appendix_data:eval}).




\textbf{Backbone Models.}
To assess both the scalability and architectural generalizability of our framework, we instantiate DLR across diverse backbones ranging from $1$B to $8$B parameters. 
For LLMs, we adopt Qwen3-4B~\cite{yang2025qwen3}, LLaMA-3.2-1B/3B~\cite{meta2024llama32}, and LLaMA-3.1-8B~\cite{grattafiori2024llama}. 
For LMMs, we use Qwen3-VL-Instruct~\cite{bai2025qwen3vl} at $2$B, $4$B, and $8$B scales~(details in App.~\ref{sec:appendix_model:backbone}).


\textbf{Baseline Methods.}
We benchmark \textbf{DLR} against representative latent reasoning methods. 
For LLMs, we compare against iCoT~\cite{deng2405explicit}~(progressive CoT removal), Coconut~\cite{hao2024coconut}~(continuous feedback), CODI~\cite{shen2025codi}~(self-distillation), CoLaR~\cite{tan2025colar}~(dynamic compression), and ReGuLaR~\cite{wang2026regular}~(variational rendering).
For LMMs, we adapt Coconut, CODI, and CoLaR to accept visual inputs and include RoT~\cite{wang2026rot}, which aligns continuous states with rendered CoT embeddings.
We report the mean $\pm$ standard deviation over five random seeds for Pass@1~(Acc.) and computational efficiency~($\#\mathrm{L}$, average token length of the reasoning chain)~(details in App.~\ref{sec:appendix_model:baseline}).

\textbf{Implementation Details.}
\label{sec:implementation}
All experiments are conducted on an $8 \times$ NVIDIA H100 node in bfloat16 mixed precision. 
We train the DeepSeek-OCR2 latent codebook via Unsloth~\cite{unsloth2023}~($3$ epochs, lr=$1\mathrm{e}{-4}$) and optimize the language models via TRL~\cite{vonwerra2020trl} through a three-stage process: (1) \emph{latent-text alignment} updates only the projectors for $3$ epochs~(lr=$1\mathrm{e}{-4}$); (2) \emph{latent SFT} updates the backbone, projectors, and codebook for $3$ epochs~(lr=$2\mathrm{e}{-5}$), training the auxiliary decoder in parallel via stop-gradient; (3) \emph{latent RL} updates the backbone and projectors using GRPO for $1$ epoch~(lr=$1\mathrm{e}{-6}$, $32$ rollouts/prompt). Further training configurations are detailed in App.~\ref{sec:appendix_implementation}.

\begin{table}[t]
\centering
\caption{Experimental results on four grade-school reasoning benchmarks with Qwen3-4B~(LLM) and Qwen3-VL-4B-Instruct~(LMM). * indicate the Pass@1 significantly outperform the second-best latent reasoning method~($p<0.05$). The results of baseline methods are from CoLaR or RoT.}
\label{tab:main_llm}
\small
\setlength{\tabcolsep}{2pt}
\resizebox{\linewidth}{!}{%
\begin{tabular}{lcccccccccc}
\toprule
 & \multicolumn{2}{c}{\textbf{GSM8K-Aug}} & \multicolumn{2}{c}{\textbf{GSM-Hard}} & \multicolumn{2}{c}{\textbf{SVAMP}} & \multicolumn{2}{c}{\textbf{MultiArith}} & \multicolumn{2}{c}{\textbf{Average}} \\
\cmidrule(lr){2-3} \cmidrule(lr){4-5} \cmidrule(lr){6-7} \cmidrule(lr){8-9} \cmidrule(lr){10-11}
\textbf{Method} & Acc. & \#L & Acc. & \#L & Acc. & \#L & Acc. & \#L & Acc. & \#L \\
\midrule
\multicolumn{11}{l}{\cellcolor{secondbg}\textit{Qwen3-4B}} \\
Direct-SFT & 27.4\textsubscript{$\pm$.4} & 0.0\textsubscript{$\pm$.0}  & 9.2\textsubscript{$\pm$.2} & 0.0\textsubscript{$\pm$.0}  & 69.4\textsubscript{$\pm$.4} & 0.0\textsubscript{$\pm$.0}  & 86.3\textsubscript{$\pm$.3} & 0.0\textsubscript{$\pm$.0}  & 48.1 & 0.0 \\
CoT-SFT    & 83.1\textsubscript{$\pm$.3} & 119.8\textsubscript{$\pm$.8}  & 52.3\textsubscript{$\pm$.3} & 189.4\textsubscript{$\pm$1.1}  & 82.9\textsubscript{$\pm$.3} & 53.7\textsubscript{$\pm$.7}  & 98.5\textsubscript{$\pm$.3} & 57.6\textsubscript{$\pm$.8}  & 79.2 & 105.1 \\\midrule
iCoT       & 13.5\textsubscript{$\pm$.2} & 0.0\textsubscript{$\pm$.0}  & 4.1\textsubscript{$\pm$.2} & 0.0\textsubscript{$\pm$.0}  & 36.9\textsubscript{$\pm$.2} & 0.0\textsubscript{$\pm$.0}  & 49.2\textsubscript{$\pm$.7} & 0.0\textsubscript{$\pm$.0}  & 25.9 & 0.0 \\
Coconut~(CoLM 2025)    & 16.9\textsubscript{$\pm$.3} & 6.0\textsubscript{$\pm$.0} & 5.4\textsubscript{$\pm$.3} & 6.0\textsubscript{$\pm$.0} & 43.6\textsubscript{$\pm$.5} & 6.0\textsubscript{$\pm$.0} & 60.3\textsubscript{$\pm$.7} & 6.0\textsubscript{$\pm$.0} & 31.6 & 6.0 \\
CODI~(EMNLP 2025)       & 7.3\textsubscript{$\pm$.5} & 6.0\textsubscript{$\pm$.0} & 2.2\textsubscript{$\pm$.2} & 6.0\textsubscript{$\pm$.0} & 11.0\textsubscript{$\pm$.6} & 6.0\textsubscript{$\pm$.0} & 18.3\textsubscript{$\pm$.8} & 6.0\textsubscript{$\pm$.0} & 9.7 & 6.0 \\
CoLaR~(NeurIPS 2025)      & 40.0\textsubscript{$\pm$.2} & 39.6\textsubscript{$\pm$.1} & 9.2\textsubscript{$\pm$.1} & 47.4\textsubscript{$\pm$.2} & 57.7\textsubscript{$\pm$.2} & 19.2\textsubscript{$\pm$.1} & 82.2\textsubscript{$\pm$.1} & 21.1\textsubscript{$\pm$.1} & 47.3 & 31.8 \\
\rowcolor{bestbg}
\textbf{DLR (SFT)} & \textbf{47.4}\textsubscript{$\pm$.4} & 6.4\textsubscript{$\pm$.3} & \textbf{13.7}\textsubscript{$\pm$.3} & 6.8\textsubscript{$\pm$.2} & \textbf{70.3}\textsubscript{$\pm$.2} & 3.6\textsubscript{$\pm$.3} & \textbf{96.1}\textsubscript{$\pm$.3} & 4.2\textsubscript{$\pm$.1} & \textbf{56.9} & 5.3 \\
\rowcolor{bestbg}
\textbf{DLR (Full)} & \textbf{62.2$^{*}$}\textsubscript{$\pm$.4} & 6.5\textsubscript{$\pm$.2} & \textbf{19.3$^{*}$}\textsubscript{$\pm$.3} & 7.0\textsubscript{$\pm$.3} & \textbf{72.3$^{*}$}\textsubscript{$\pm$.2} & 3.8\textsubscript{$\pm$.2} & \textbf{97.8$^{*}$}\textsubscript{$\pm$.3} & 4.3\textsubscript{$\pm$.3} & \textbf{62.9} & 5.4 \\
\midrule
\multicolumn{11}{l}{\cellcolor{secondbg}\textit{Qwen3-VL-4B-Instruct}} \\
Direct-SFT & 26.2\textsubscript{$\pm$.3} & 0.0\textsubscript{$\pm$.0}  & 9.5\textsubscript{$\pm$.1} & 0.0\textsubscript{$\pm$.0}  & 70.0\textsubscript{$\pm$.3} & 0.0\textsubscript{$\pm$.0}  & 85.6\textsubscript{$\pm$.4} & 0.0\textsubscript{$\pm$.0}  & 47.8 & 0.0 \\
CoT-SFT    & 81.2\textsubscript{$\pm$.4} & 127.3\textsubscript{$\pm$1.0}  & 53.4\textsubscript{$\pm$.3} & 191.1\textsubscript{$\pm$1.6}  & 84.3\textsubscript{$\pm$.3} & 55.9\textsubscript{$\pm$.6}  & 98.3\textsubscript{$\pm$.4} & 59.1\textsubscript{$\pm$.6}  & 79.3 & 108.4 \\\midrule
Coconut~(CoLM 2025) & 16.1\textsubscript{$\pm$.4} & 6.0\textsubscript{$\pm$.0} & 5.2\textsubscript{$\pm$.3} & 6.0\textsubscript{$\pm$.0} & 42.0\textsubscript{$\pm$.3} & 6.0\textsubscript{$\pm$.0} & 58.9\textsubscript{$\pm$.2} & 6.0\textsubscript{$\pm$.0} & 30.5 & 6.0 \\
CODI~(EMNLP 2025)    & 7.1\textsubscript{$\pm$.3} & 6.0\textsubscript{$\pm$.0} & 2.0\textsubscript{$\pm$.4} & 6.0\textsubscript{$\pm$.0} & 10.7\textsubscript{$\pm$.1} & 6.0\textsubscript{$\pm$.0} & 17.8\textsubscript{$\pm$.2} & 6.0\textsubscript{$\pm$.0} & 9.4 & 6.0 \\
CoLaR~(NeurIPS 2025)   & 39.2\textsubscript{$\pm$.2} & 38.8\textsubscript{$\pm$.5} & 8.2\textsubscript{$\pm$.4} & 47.8\textsubscript{$\pm$.7} & 52.0\textsubscript{$\pm$.4} & 18.4\textsubscript{$\pm$.6} & 81.0\textsubscript{$\pm$.3} & 22.1\textsubscript{$\pm$.3} & 45.1 & 31.8 \\
RoT~(ACL 2026)     & 37.8\textsubscript{$\pm$.3} & 32.0\textsubscript{$\pm$.0} & 14.1\textsubscript{$\pm$.2} & 32.0\textsubscript{$\pm$.0} & 72.7\textsubscript{$\pm$.4} & 32.0\textsubscript{$\pm$.0} & 97.2\textsubscript{$\pm$.4} & 32.0\textsubscript{$\pm$.0} & 55.4 & 32.0 \\
\rowcolor{bestbg}
\textbf{DLR (SFT)} & \textbf{48.0}\textsubscript{$\pm$.4} & 6.3\textsubscript{$\pm$.4} & \textbf{14.7}\textsubscript{$\pm$.3} & 6.8\textsubscript{$\pm$.5} & \textbf{68.7}\textsubscript{$\pm$.2} & 3.6\textsubscript{$\pm$.1} & \textbf{97.8}\textsubscript{$\pm$.3} & 4.2\textsubscript{$\pm$.2} & \textbf{57.3} & 5.2 \\
\rowcolor{bestbg}
\textbf{DLR (Full)} & \textbf{63.3$^{*}$}\textsubscript{$\pm$.3} & 6.5\textsubscript{$\pm$.1} & \textbf{18.3$^{*}$}\textsubscript{$\pm$.4} & 7.0\textsubscript{$\pm$.6} & \textbf{72.7}\textsubscript{$\pm$.3} & 3.8\textsubscript{$\pm$.4} & \textbf{98.3$^{*}$}\textsubscript{$\pm$.2} & 4.3\textsubscript{$\pm$.3} & \textbf{63.2} & 5.4 \\
\bottomrule
\end{tabular}}
\end{table}

\subsection{Main Results}
\label{sec:main_results}

We evaluate \textbf{DLR} along two complementary axes. 
First, we benchmark performance and out-of-domain~(OOD) generalization on grade-school reasoning tasks~(GSM8K-Aug, GSM-Hard, SVAMP, MultiArith) using 4B LLMs and LMMs~(Tab.~\ref{tab:main_llm}).
Second, we assess its scalability and reasoning capability on the high-school-level MATH-500 benchmark across three model scales~(Tab.~\ref{tab:main_math}).
\textbf{Additional experiments on different LLM and LMM backbones are detailed in App.~\ref{sec:appendix_main}.}

\textbf{Results on Grade-School-Level Reasoning and OOD Generalization.}
Tab.~\ref{tab:main_llm} compares \textbf{DLR} against baselines on both LLM and LMM backbones.
On the in-domain GSM8K-Aug dataset, \textbf{DLR} achieves the highest accuracy among latent methods while \textbf{reducing the reasoning length} to $\sim\!6$ tokens~(versus $119$ in CoT-SFT and $32$ in RoT).
On OOD benchmarks like GSM-Hard and SVAMP, continuous latent baselines suffer from severe error accumulation and representation drift.
In contrast, \textbf{DLR} demonstrates \textbf{strong out-of-domain generalization}, significantly outperforming the second-best baseline~($p<0.05$) while maintaining a superior compression rate.
This confirms that our discrete codebook provides stable symbolic anchors that prevent latent drift, regardless of whether the underlying backbone is an LLM or LMM.

\textbf{Results on High-School-Level Scalability.}
To examine scalability on harder reasoning tasks, Tab.~\ref{tab:main_math} reports results on MATH-500 across three LMM scales~(2B, 4B, 8B).
As capacity increases, \textbf{DLR} exhibits \textbf{consistent scaling behavior}.
While the rendered-CoT baseline~(RoT) shows diminishing returns from 4B to 8B~(+4.6 points), \textbf{DLR~(Full)} consistently scales, achieving a +9.4 point gain to reach $54.0\%$ accuracy on the 8B backbone.
Comparing \textbf{DLR~(SFT)} to \textbf{DLR~(Full)} highlights a \textbf{performance boost from latent RL}, confirming that discrete tokens reliably support reward-driven policy optimization.
Moreover, \textbf{DLR} achieves this using fewer than $30$ latent tokens, offering much higher compression than RoT~($64$) and explicit CoT~($\sim\!300$), demonstrating that our discrete vocabulary efficiently unlocks semantic capacity without acting as an information bottleneck.

\begin{table}[t]
\centering
\caption{Experimental results on MATH-500 with Qwen3-VL at three scales. We compare \textbf{DLR} against SFT-w/o CoT, explicit SFT-CoT, and the rendered-CoT baseline RoT.}
\label{tab:main_math}
\small
\setlength{\tabcolsep}{4pt}
\resizebox{0.8\linewidth}{!}{%
\begin{tabular}{lcccccc}
\toprule
 & \multicolumn{2}{c}{\textbf{Qwen3-VL-2B}} & \multicolumn{2}{c}{\textbf{Qwen3-VL-4B}} & \multicolumn{2}{c}{\textbf{Qwen3-VL-8B}} \\
\cmidrule(lr){2-3} \cmidrule(lr){4-5} \cmidrule(lr){6-7}
\textbf{Method} & Acc. & \#L & Acc. & \#L & Acc. & \#L \\
\midrule
Direct-SFT  & 20.8\textsubscript{$\pm$.2} & 0.0\textsubscript{$\pm$.0}   & 29.4\textsubscript{$\pm$.3} & 0.0\textsubscript{$\pm$.0}   & 33.5\textsubscript{$\pm$.5} & 0.0\textsubscript{$\pm$.0} \\
CoT-SFT     & 29.2\textsubscript{$\pm$.3} & 324.5\textsubscript{$\pm$2.6} & 55.8\textsubscript{$\pm$.4} & 291.5\textsubscript{$\pm$1.9} & 69.4\textsubscript{$\pm$.3} & 310.6\textsubscript{$\pm$1.7} \\
RoT         & 24.0\textsubscript{$\pm$.2} & 64.0\textsubscript{$\pm$.0} & 33.2\textsubscript{$\pm$.4} & 64.0\textsubscript{$\pm$.0} & 37.8\textsubscript{$\pm$.3} & 64.0\textsubscript{$\pm$.0} \\
\rowcolor{bestbg}
\textbf{DLR (SFT)} & \textbf{29.4}\textsubscript{$\pm$.3} & 26.7\textsubscript{$\pm$.8} & \textbf{35.4}\textsubscript{$\pm$.3} & 27.9\textsubscript{$\pm$.7} & \textbf{39.8}\textsubscript{$\pm$.4} & 27.2\textsubscript{$\pm$.5} \\
\rowcolor{bestbg}
\textbf{DLR (Full)} & \textbf{34.0$^{*}$}\textsubscript{$\pm$.3} & 28.9\textsubscript{$\pm$.6} & \textbf{44.6$^{*}$}\textsubscript{$\pm$.4} & 28.4\textsubscript{$\pm$.5} & \textbf{54.0$^{*}$}\textsubscript{$\pm$.3} & 29.5\textsubscript{$\pm$.6} \\
\bottomrule
\end{tabular}}
\end{table}

\begin{table}[t]
\centering
\caption{Ablation on training-stage composition with Qwen3-VL-4B-Instruct. \cmark indicate which components are enabled: Stage~1 (latent-text alignment), Stage~2 (latent SFT), Stage~3 (latent RL), and the process alignment reward, $r_i^\mathrm{proc}$ (Proc.) inside Stage~3. The last row is our full configuration.}
\label{tab:ablation_stage}
\small
\setlength{\tabcolsep}{4pt}
\resizebox{\linewidth}{!}{%
\begin{tabular}{lcccccccccccc}
\toprule
 & \multicolumn{4}{c}{\textbf{Components}} & \multicolumn{2}{c}{\textbf{GSM8K-Aug}} & \multicolumn{2}{c}{\textbf{GSM-Hard}} & \multicolumn{2}{c}{\textbf{SVAMP}} & \multicolumn{2}{c}{\textbf{MultiArith}} \\
\cmidrule(lr){2-5} \cmidrule(lr){6-7} \cmidrule(lr){8-9} \cmidrule(lr){10-11} \cmidrule(lr){12-13}
\textbf{Method} & \textbf{S1} & \textbf{S2} & \textbf{S3} & \textbf{Proc.} & Acc. & \#L & Acc. & \#L & Acc. & \#L & Acc. & \#L \\
\midrule
--\,Align            & \xmark & \cmark & \cmark & \cmark & 49.6\textsubscript{$\pm$.2} & 6.4 & 14.3\textsubscript{$\pm$.3} & 6.6 & 65.3\textsubscript{$\pm$.2} & 3.7 & 93.9\textsubscript{$\pm$.3} & 4.1 \\
--\,SFT              & \cmark & \xmark & \cmark & \cmark & 21.4\textsubscript{$\pm$.3} & 8.5 & 3.2\textsubscript{$\pm$.5} & 11.3 & 42.3\textsubscript{$\pm$.4} & 9.7 & 49.4\textsubscript{$\pm$.6} & 8.4 \\
--\,RL               & \cmark & \cmark & \xmark & \cmark & 48.0\textsubscript{$\pm$.4} & 6.3 & 14.7\textsubscript{$\pm$.3} & 6.8 & 68.7\textsubscript{$\pm$.2} & 3.6 & 97.8\textsubscript{$\pm$.3} & 4.2 \\
--\,Proc.\,Rwd.      & \cmark & \cmark & \cmark & \xmark & 59.6\textsubscript{$\pm$.2} & 6.6 & 16.2\textsubscript{$\pm$.3} & 6.7 & 72.3\textsubscript{$\pm$.1} & 3.6 & 98.3\textsubscript{$\pm$.4} & 4.3 \\
\rowcolor{bestbg}
\textbf{DLR (Full)}  & \cmark & \cmark & \cmark & \cmark & \textbf{63.3}\textsubscript{$\pm$.3} & 6.5 & \textbf{18.3}\textsubscript{$\pm$.4} & 7.0 & \textbf{72.7}\textsubscript{$\pm$.3} & 3.8 & \textbf{98.3}\textsubscript{$\pm$.2} & 4.3 \\
\bottomrule
\end{tabular}}
\end{table}

\subsection{Ablation Study}
\label{sec:ablation}

All ablations are conducted on Qwen3-VL-4B-Instruct and evaluated on the four grade-school benchmarks.
\textbf{Further ablations on the latent codebook and latent decoding are in App.~\ref{sec:appendix_ablation}.}

\begin{figure*}[t]
  \centering
  \includegraphics[width=\textwidth]{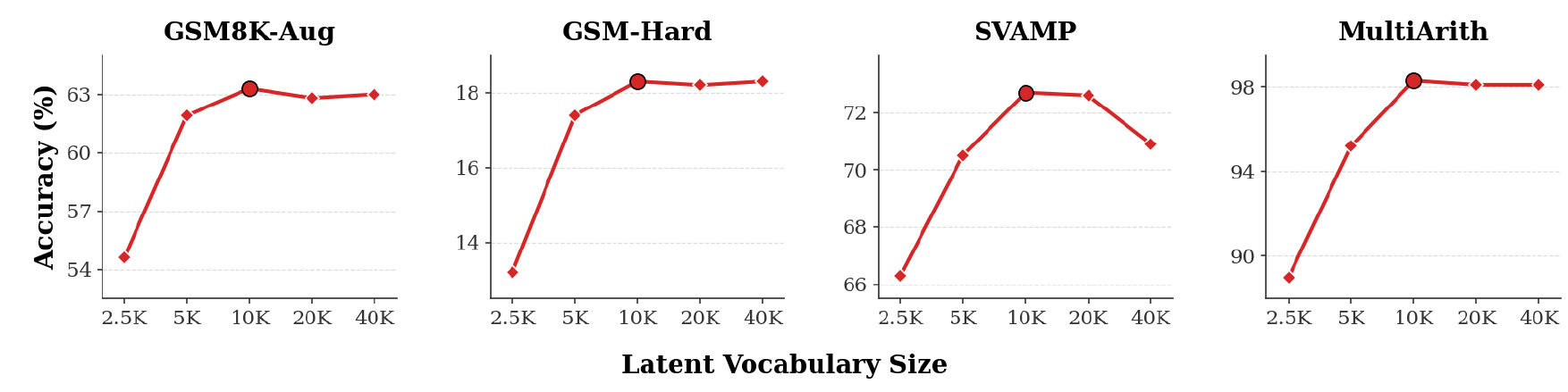}
  \caption{Effect of latent vocabulary size ($K$) on reasoning performance.}
  \label{fig:ablation_vocab}
\end{figure*}

\textbf{Latent Vocabulary Capacity.}
We first study the effect of the latent codebook size by varying $K \in \{2.5\mathrm{K}, 5\mathrm{K}, 10\mathrm{K}, 20\mathrm{K}, 40\mathrm{K}\}$~(Fig.~\ref{fig:ablation_vocab}).
Performance follows a clear two-phase trend: from $K=2.5\mathrm{K}$ to $10\mathrm{K}$, accuracy steadily improves across all four benchmarks as the vocabulary acquires sufficient capacity to capture diverse reasoning semantics.
However, from $K=10\mathrm{K}$ to $40\mathrm{K}$, performance plateaus and occasionally degrades slightly, suggesting that an excessively large codebook leads to under-utilized codes and limits generalization.
Thus, we set $K=10\mathrm{K}$ to balance representation capacity and training stability.

\textbf{Training-Stage Composition.}
Tab.~\ref{tab:ablation_stage} analyzes the contribution of our three-stage process and the specific reward design.
Removing Stage-1 Alignment~($-$\,Align) forces the latent tokens to co-train with the backbone from scratch, failing to ground them semantically and causing a major performance drop.
Skipping Stage-2 SFT~($-$\,SFT) causes a performance collapse, indicating that RL cannot learn complex latent distributions from sparse outcome signals alone.
Furthermore, while the base Stage-3 RL provides clear refinement over SFT~($-$\,RL), disabling the \emph{Process Alignment Reward}~($-$\,Proc.\,Rwd.) strictly degrades accuracy.
This demonstrates that the process reward is critical for keeping the latent trajectory semantically structured during policy optimization.

\subsection{Interpretability through Case Study}
\begin{figure*}[t]
  \centering
  \includegraphics[width=\textwidth]{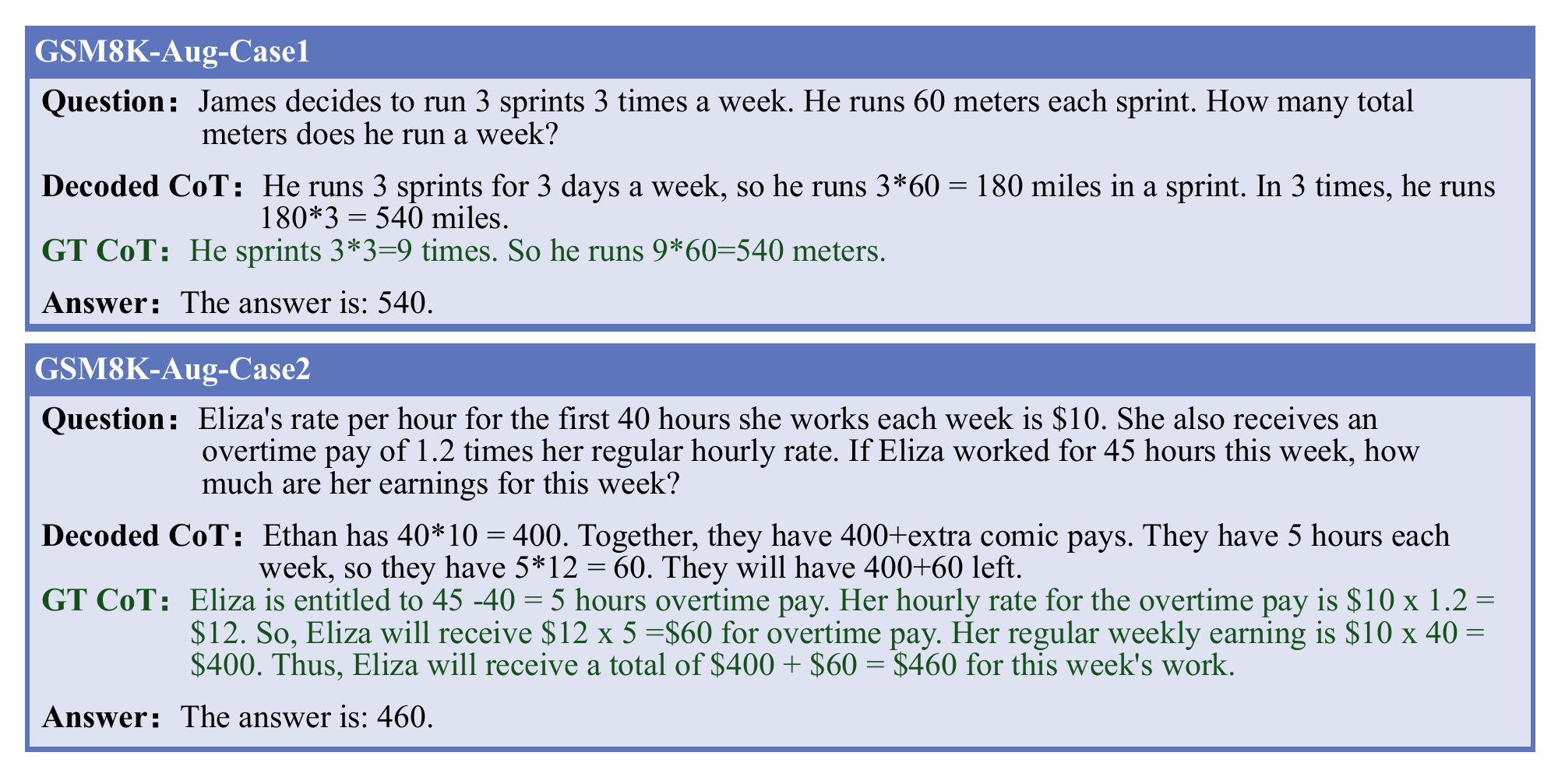}
  \caption{Case study of latent-to-text decoding. Best viewed by zoom in.}
  \label{fig:case_decoding}
\end{figure*}

Fig.~\ref{fig:case_decoding} illustrates how our auxiliary decoder translates latent trajectories into interpretable steps.
The decoded chains of thought capture the correct mathematical logic and consistently yield accurate final answers.
While the decoder occasionally hallucinates trivial semantic details during reconstruction, such as substituting incorrect names~(\emph{e.g.}, ``Ethan'' for ``Eliza'') or units~(\emph{e.g.}, ``miles'' for ``meters''), these minor artifacts are inconsequential and do not disrupt the underlying causal reasoning of the latent states.
This validates the strong interpretability of our discrete latent method.
Additional case studies are provided in App.~\ref{sec:appendix_interpretablity}.

\section{Conclusion}
In this paper, we resolve the misalignment between continuous latent states and discrete autoregressive supervision by proposing \textbf{Discrete Latent Reasoning~(DLR)}.
\textbf{DLR} represents intermediate reasoning steps as explicit discrete tokens using a visual-semantic codebook constructed from rendered chains of thought.
To integrate these tokens, we design a three-stage training process with dual-branch supervised fine-tuning and a novel Process Alignment Reward, maintaining both training stability and trajectory interpretability.
Extensive experiments across diverse LLM and LMM architectures confirm that \textbf{DLR} achieves superior accuracy, strong out-of-domain generalization, and consistent scaling on mathematical benchmarks.
By dynamically compressing the reasoning length while retaining semantic interpretability, \textbf{DLR} offers an efficient, controllable, and scalable approach that advances the development of latent reasoning models.

\section*{Acknowledgements}

This work was supported by the National Natural Science Foundation
of China (Grant No. 62471287).

\bibliographystyle{plainnat}
\bibliography{references}

\clearpage 
\appendix
This appendix provides additional materials to supplement the main submission.
Section~\ref{sec:appendix_data} details the data curation procedures, including the chain-of-thought rendering process and the specific formatting protocols for both codebook and language model training.
Section~\ref{sec:appendix_model} provides further details on the backbone architectures~(both LLMs and LMMs) and elaborates on the mechanisms of the baseline latent reasoning methods used in our evaluation.
Section~\ref{sec:appendix_implementation} outlines the implementation details, including hardware setups, software frameworks, and hyperparameters for latent codebook construction and all three training stages of the latent language model.
Section~\ref{sec:appendix_main} presents additional comparative results on LLaMA-3.2-1B and QwenVL3-2B-Instruct to verify the effectiveness of the DLR method.
Section~\ref{sec:appendix_ablation} reports further ablation studies that analyze the impact of various hyperparameter choices, codebook sizes, and reward configurations on model performance.
Section~\ref{sec:appendix_interpretablity} provides extended interpretability analyses, showcasing qualitative examples of latent trajectories translated into explicit reasoning steps by our auxiliary OCR decoder.
Section~\ref{sec:appendix_limitation} discusses the current limitations of our approach and explores the broader societal impacts of deploying efficient, highly capable reasoning models.

\section{Details of Dataset}\label{sec:appendix_data}

\subsection{Details of Training Data}\label{sec:appendix_data:train}

As outlined in the main text, \textbf{DLR} involves two distinct training stages that require separate data curation procedures: (i) the latent codebook, which is trained on rendered CoT images, and (ii) the latent language model, which is trained over mixed sequences of text and latent tokens. Below, we detail the data processing and formatting protocols for each stage.

\subsubsection{Latent Codebook Training Data}
For codebook training, we aggregate a corpus of approximately $1$M reasoning traces covering a broad spectrum of mathematical difficulty. 
Specifically, we include the full GSM8K-Aug training split~($385$K)~\cite{deng2023implicit}, the MATH training split~($12$K)~\cite{hendrycks2021math} obtained by excluding MATH-500 from the full MATH dataset, and a stratified subsample of $603$K instances from MathX-5M~\cite{modotte2026mathx5m}.
We extract the chain-of-thought segment from each instance and discard natural-language preambles. 
To ensure that every trace fits into a single rendered canvas, we retain only samples whose CoT length is at most $2,048$ text tokens, measured with the DeepSeek-OCR2 tokenizer~\cite{wei2026deepseekocr2}. 
Each retained CoT is rendered into a square RGB image via a Python pipeline, with the font size Gaussian-sampled in $[15, 20]$\,pt and the canvas side length adaptively chosen to fit the rendered content. 
The resulting side length is clipped to the range $[64, 1024]$ pixels and resized to the nearest endpoint when it falls outside this range. 
Under the DeepEncoder V2~\cite{wei2026deepseekocr2} tokenization scheme, the number of visual tokens produced from a side length $L$ follows $N_{\mathrm{vis}} = \left( \frac{L}{64} \right)^2,$ so that a $1024 \times 1024$ canvas yields exactly $256$ visual tokens. 
In the upper-bound setting, this corresponds to a $\sim\!8\times$ compression ratio relative to the $2048$-token CoT budget, consistent with the render-compression rationale of our approach.
The codebook is then trained with the rendered image as input and the original textual CoT as decoding supervision for the auxiliary decoder $\mathbf{D}_{\mathrm{ocr}}$.

\subsubsection{Latent Language Model Training Data}
For latent language model training, we construct two configurations.
The \emph{base} configuration uses only the GSM8K-Aug split~\cite{deng2405explicit}, reusing the $385$K rendered images and their codebook-derived latent token sequences.
Within this split, the full set is used for Stage~1 latent-text alignment, the first $300$K instances are used for Stage~2 supervised fine-tuning~(SFT), and the remaining $85$K instances are used for Stage~3 reinforcement learning~(RL).
During SFT, each sample is formatted as the triple $[x;\, z_{1:n};\, a_{1:M}]$, where $z_{1:n}$ is obtained from the pretrained codebook; during RL, only the problem $x$ is fed to the model and the latent trajectory is generated autoregressively.
The \emph{scaled} configuration, designed to assess generalization to harder mathematical reasoning, uses the MATH training split~\cite{hendrycks2021math} of $\sim\!12$K instances, allocating the full set to alignment, $7.5$K to SFT, and $4.5$K to RL following the same protocol.

\subsection{Details of Evaluation Data}\label{sec:appendix_data:eval}

To evaluate the arithmetic, out-of-domain, and advanced mathematical reasoning capabilities of our models, we conduct zero-shot evaluations across five standard benchmarks.
No ground-truth explicit chains of thought are provided at test time.
The detailed statistics and task descriptions for each evaluation dataset are as follows:

\begin{itemize}
    \item \textbf{GSM8K}~\cite{cobbe2021training}: Consists of $1{,}319$ test problems, forming a widely used benchmark of grade-school math word problems that require multi-step arithmetic reasoning.
    \item \textbf{GSM-Hard}~\cite{gao2023pal}: Contains $1{,}319$ test problems constructed by replacing numerical values in the original GSM8K test set with larger numbers, serving as a reliable metric to test whether the model relies on superficial heuristics or mathematical logic.
    \item \textbf{SVAMP}~\cite{patel2021nlp}: Comprises $1{,}000$ test problems, serving as a challenge set created by applying structural and linguistic variations to existing word problems to evaluate a model's stability against varying descriptions.
    \item \textbf{MultiArith}~\cite{roy2015solving}: Includes $600$ multi-step arithmetic problems. We use the full dataset as a standard out-of-domain~(OOD) evaluation benchmark to measure the generalization capabilities of models trained primarily on GSM8K.
    \item \textbf{MATH-500}~\cite{hendrycks2021math,lightman2024lets}: Consists of $500$ highly challenging test problems, forming a curated subset of the MATH dataset that features competition-level mathematics problems across advanced subjects such as algebra, geometry, and number theory.
\end{itemize}

\section{Details of Comparison Baselines}\label{sec:appendix_model}

\subsection{Details of Backbone LLMs and LMMs}\label{sec:appendix_model:backbone}

As outlined in the main text, we instantiate our method across multiple backbone architectures to evaluate its generalizability and scalability.
Specifically, we adopt the following model series:

\begin{itemize}
    \item \textbf{LLaMA Series (3.2-1B/3B and 3.1-8B)}~\cite{meta2024llama32,grattafiori2024llama}: As a representative open-weight LLM family developed by Meta, LLaMA offers highly optimized language models. 
    We utilize LLaMA-3.2-1B/3B and LLaMA-3.1-8B to assess how discrete latent reasoning scales with model capacity within pure text domains.
    \item \textbf{Qwen3 Dense Text Models (4B)}~\cite{yang2025qwen3}: Developed by Alibaba Cloud, the Qwen3 series provides state-of-the-art dense language models.
    We select the Qwen3-4B model as an alternative mid-scale text-only backbone, allowing us to evaluate the adaptability of our discrete latent method across different pre-training recipes and tokenizer designs.
    \item \textbf{Qwen3-VL-Instruct Series (2B, 4B, 8B)}~\cite{bai2025qwen3vl}: To verify that our approach is agnostic to modality, we extend our evaluation to the large multimodal model~(LMM) domain using the Qwen3-VL-Instruct family.
    Evaluating these native vision-language models at $2$B, $4$B, and $8$B scales enables us to analyze the scalability of latent reasoning when processing visual representations and rendered reasoning traces.
\end{itemize}

\subsection{Details of Compared Latent Reasoning Methods}\label{sec:appendix_model:baseline}

To evaluate the effectiveness of our proposed method, we compare \textbf{DLR} against a representative set of state-of-the-art latent reasoning baselines, grouped by the modality of their underlying backbone:

\noindent \textbf{Large Language Model Baselines:}
\begin{itemize}
    \item \textbf{iCoT}~\cite{deng2405explicit}: This method starts from a model trained on explicit chain-of-thought and progressively removes intermediate reasoning tokens during fine-tuning. This curriculum forces the model to internalize the reasoning process directly within its hidden states.
    \item \textbf{Coconut}~\cite{hao2024coconut}: A pioneering approach that operates in a purely continuous latent space. It feeds the last-layer hidden state back as the next input embedding, bypassing the LM head and tokenization under a multi-stage curriculum that gradually replaces textual steps with continuous thoughts.
    \item \textbf{CODI}~\cite{shen2025codi}: This method casts implicit reasoning as a self-distillation problem. A single shared model is jointly trained under explicit and implicit CoT objectives, aligning the hidden activation at the answer token between the two modes so that continuous latent states inherit the semantics of explicit rationales.
    \item \textbf{CoLaR}~\cite{tan2025colar}: This method compresses groups of explicit reasoning-token embeddings into dense latent tokens with a prompt-controllable compression factor, allowing the model to dynamically adjust its reasoning ``speed'' and computational cost at inference time.
    \item \textbf{ReGuLaR}~\cite{wang2026regular}: ReGuLaR situates latent reasoning within a variational auto-encoding architecture. Each latent state is sampled from a posterior, and its prior is regularized by dense visual-semantic features obtained by rendering the corresponding CoT segment and encoding it with a frozen DeepSeek-OCR encoder.
\end{itemize}

\noindent \textbf{Large Multimodal Model Baselines:}
\begin{itemize}
    \item \textbf{Render-of-Thought (RoT)}~\cite{wang2026rot}: As the most directly comparable visual-rendering baseline to our method, RoT renders each CoT step into a single-line image. It then trains the model to produce continuous hidden states that align with the visual embeddings extracted by a frozen VLM encoder, effectively using rendered text as an external semantic anchor to supervise latent reasoning.
\end{itemize}

\begin{table}[t]
\centering
\caption{Detailed per-stage training configurations of \textbf{DLR}. ``Trainable Modules'' lists the parameters updated in each stage, while the rest are frozen. The symbol ``$-$'' indicates not applicable.}
\label{tab:implementation_detailed}
\small
\resizebox{\textwidth}{!}{
\begin{tabular}{lcccc}
\toprule
\textbf{Hyperparameter} & \textbf{Codebook} & \textbf{Stage 1: Align} & \textbf{Stage 2: SFT} & \textbf{Stage 3: RL (GRPO)} \\
\midrule
Software Framework       & Unsloth & TRL & TRL & TRL \\
DeepSpeed Stage          & - & Zero-2 & Zero-2 & Zero-2 \\
Optimizer                & AdamW & AdamW & AdamW & AdamW \\
Peak Learning Rate       & $1\mathrm{e}{-4}$ & $1\mathrm{e}{-4}$ & $2\mathrm{e}{-5}$ & $1\mathrm{e}{-6}$ \\
LR Schedule              & Cosine Decay & Cosine Decay & Cosine Decay & Cosine Decay \\
Warmup Ratio             & $0.03$ & $0.03$ & $0.03$ & $0.03$ \\
Weight Decay             & $1\mathrm{e}{-2}$ & $1\mathrm{e}{-2}$ & $1\mathrm{e}{-2}$ & $1\mathrm{e}{-2}$ \\
Epochs                   & $3$ & $3$ & $3$ & $1$ \\
Max Sequence Length      & 256 & 256 & 256 & 256 \\
Per-GPU Batch Size       & $8$ & $8$ & $8$ & $8$ \\
Global Batch Size        & $64$ & $64$ & $64$ & $64$ \\
\midrule
Rollout Size ($G$)       & $-$ & $-$ & $-$ & $32$ \\
PPO Inner Epochs         & $-$ & $-$ & $-$ & $4$ \\
Sampling Temperature     & $-$ & $-$ & $-$ & $1.0$ \\
KL Penalty ($\beta_{\mathrm{KL}}$) & $-$ & $-$ & $-$ & $1\mathrm{e}{-3}$ \\
Reward Weights           & $-$ & $-$ & $-$ & $\lambda_{\mathrm{proc}}=\mathrm{0.25}, \lambda_{\mathrm{fmt}}=\mathrm{0.1}$ \\
\midrule
\textbf{Trainable Modules} & Enc. + $C$ + $\mathbf{D}_{\mathrm{ocr}}$ & $P_{\mathrm{in}}, P_{\mathrm{out}}$ & LM + $C$ + $P_{*}$ \newline (+ $\mathbf{D}_{\mathrm{ocr}}$) & LM + $P_{*}$ \\
\textbf{FLOPS} & 780P & 208P & 325P & 1365P \\
\bottomrule
\end{tabular}
}
\end{table}

\section{Detailed Implementation Configurations}\label{sec:appendix_implementation}

This section provides specifics regarding the hardware setups, software libraries, and hyperparameters used for both the latent codebook training and the three-stage latent language model training.
A complete summary of the per-stage training configurations is provided in Table~\ref{tab:implementation_detailed}.

\subsection{Frameworks and Hardware}
All experiments are conducted on an $8 \times$ NVIDIA H100~(80GB) node.
To maximize throughput and memory efficiency, we employ mixed-precision \texttt{bfloat16} training across all stages. 
For the latent codebook, we leverage the Unsloth library~\cite{unsloth2023} to optimize the DeepSeek-OCR2 architecture~\cite{wei2026deepseekocr2}.
Its DeepEncoder V2 consists of a SAM-based~\cite{kirillov2023segment} local encoder, a $16\times$ convolutional compressor, and a Qwen2~\cite{yang2024qwen2} causal-flow encoder, while the decoder $\mathbf{D}_{\mathrm{ocr}}$ follows the DeepSeek-LLM~backbone. 
For the latent language model training, we utilize the TRL~(Transformer Reinforcement Learning) library~\cite{vonwerra2020trl} integrated with DeepSpeed Zero-2~\cite{rasley2020deepspeed} to distribute optimizer states and gradients.
Furthermore, FlashAttention-2~\cite{daoflashattention} is enabled by default to accelerate self-attention computation for long multimodal sequences.

\subsection{Latent Codebook Training Configurations}
In accordance with the formulations in the main text, the entire codebook stack, including the visual encoder, the causal-flow module, the discrete codebook $C$, and the decoder $\mathbf{D}_{\mathrm{ocr}}$, is trained end-to-end.
The latent codebook has a vocabulary size of $K=\mathrm{10,000}$ with a code dimension of $d_c=\mathrm{896}$. 
The vector-quantization loss weight is set to $\lambda_{\mathrm{VQ}}=\mathrm{0.25}$, and the commitment loss weight is $\beta=\mathrm{0.1}$.

We optimize the model using the AdamW~\cite{loshchilovdecoupled} optimizer with $\beta_1=\mathrm{0.9}$, $\beta_2=\mathrm{0.999}$, and a weight decay of $\mathrm{0.01}$. 
The training spans $3$ epochs on the $1$M rendered corpus with a peak learning rate of $1\mathrm{e}{-4}$ and a global batch size of $\mathrm{64}$.
To prevent early codebook collapse, we employ a two-branch curriculum where the continuous-branch annealing factor $\alpha(e)$ is linearly decayed from $1$ to $0$ over the course of the first epoch.

\subsection{Latent Language Model Training Configurations}
The language models are trained with a maximum sequence length of $L_{\max}=\mathrm{256}$ text and latent tokens. 
The learning rates for all stages follow a cosine-annealing schedule with a $0.03$ warmup ratio. 
The specific setups for the three stages are as follows:

\begin{itemize}
    \item \textbf{Stage 1 (Latent-Text Alignment):} Only the input/output projectors $(P_{\mathrm{in}}, P_{\mathrm{out}})$ are optimized, while the LLM backbone is completely frozen.
    The projectors are implemented as single linear layers mapping between the codebook dimension $d_c$ and the backbone hidden dimension $d_{\mathrm{lm}}$~(e.g., $d_{\mathrm{lm}}=4096$ for Qwen3-VL-8B). 
    This stage runs for $3$ epochs with a learning rate of $1\mathrm{e}{-4}$ and a global batch size of $\mathrm{64}$.
    \item \textbf{Stage 2 (Latent Supervised Fine-Tuning):} The LLM/LMM backbone, the codebook embeddings, and the projectors are jointly updated.
    The auxiliary decoder $\mathbf{D}_{\mathrm{ocr}}$ is trained in parallel, but its gradients are stopped at the backbone and codebook interface. This stage runs for $3$ epochs with a per-GPU batch size of $8$~(resulting in a global batch size of $\mathrm{64}$), a learning rate of $2\mathrm{e}{-5}$, and an AdamW weight decay of $\mathrm{0.01}$.
    \item \textbf{Stage 3 (Latent Reinforcement Learning):} The codebook and $\mathbf{D}_{\mathrm{ocr}}$ are strictly frozen, while the backbone and projectors are updated via GRPO. This stage is trained for $1$ epoch with a learning rate of $1\mathrm{e}{-6}$. We sample $G=32$ rollouts per prompt using a temperature of $1.0$. The GRPO algorithm performs $4$ PPO inner epochs per batch. The reward coefficients are set to $\lambda_{\mathrm{proc}}=\mathrm{0.25}$ for the process alignment reward and $\lambda_{\mathrm{fmt}}=\mathrm{0.1}$ for the format verification reward, while the KL-divergence penalty coefficient against the reference policy is set to $\beta_{\mathrm{KL}}=\mathrm{0.001}$.
\end{itemize}

\section{Additional Main Results}\label{sec:appendix_main}

To expand the evaluation of our proposed method, this section presents extended experimental results on smaller model scales and detailed scalability analyses.

\subsection{Extended Baseline Comparisons at 1B and 2B Scales}
Table~\ref{tab:main_lmm} reports the detailed performance of \textbf{DLR} against all baselines on the lightweight LLaMA-3.2-1B and Qwen3-VL-2B-Instruct architectures.
The results confirm that the advantages of discrete latent reasoning persist even in capacity-constrained regimes.
On the \textbf{LLaMA-3.2-1B} backbone, \textbf{DLR} achieves $51.3\%$ average accuracy across the four grade-school benchmarks.
It clearly outperforms the strongest continuous-latent baseline, ReGuLaR~($45.6\%$), which also leverages rendered CoT but relies on a continuous variational prior.
Furthermore, \textbf{DLR} maintains a compact latent footprint~(averaging $\sim\!5.2$ tokens), offering a favorable trade-off between computational efficiency and reasoning accuracy compared to explicit CoT-SFT~($113.0$ tokens).

This advantage transfers directly to the \textbf{Qwen3-VL-2B-Instruct} multimodal backbone.
\textbf{DLR} outperforms the visually aligned continuous baseline, RoT~($37.0\%$ average accuracy), establishing a new state-of-the-art for lightweight latent reasoning models.
Specifically, RoT requires a fixed length of $32$ continuous tokens per reasoning step, whereas \textbf{DLR} achieves higher accuracy while using a dynamically generated, compressed discrete latent sequence.
This confirms that explicitly discretizing the latent space provides a stronger semantic anchor for lightweight models than purely continuous visual alignment.

\subsection{Scalability Analysis Across Model Sizes}
\begin{figure*}[t]
  \centering
  \includegraphics[width=\textwidth]{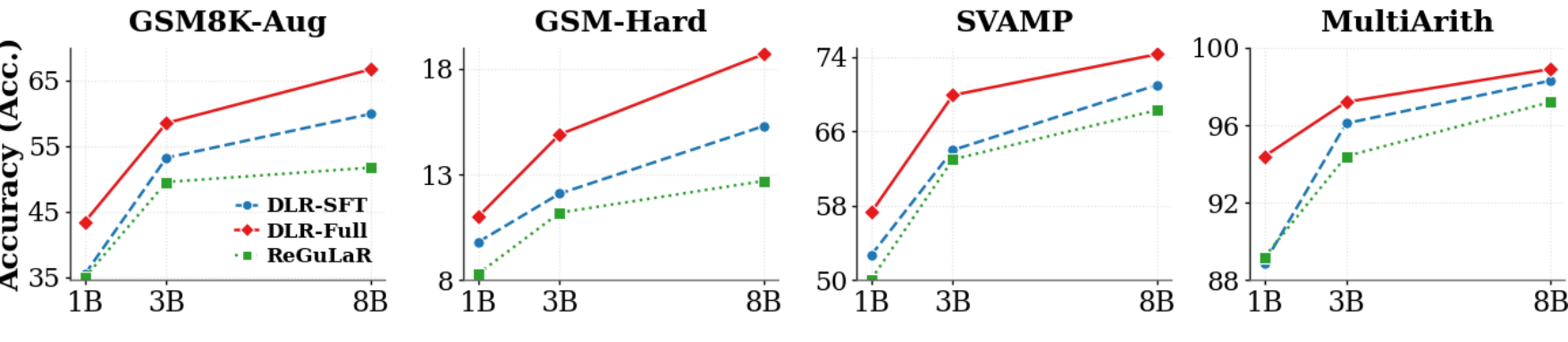}
  \caption{Scalability analysis across varying model sizes.}
  \label{fig:scale_llama}
\end{figure*}

To investigate how different latent reasoning methods benefit from increased model capacity, we analyze the scalability across 1B, 3B, and 8B parameter scales on the four grade-school-level benchmarks~(Fig.~\ref{fig:scale_llama}).

As illustrated by the scaling trajectories, while continuous latent methods such as Coconut and CoLaR show initial improvements from 1B to 3B, they face diminishing returns and plateau at the 8B scale.
ReGuLaR achieves higher baseline performance but similarly suffers from scaling bottlenecks, particularly on out-of-domain tasks like GSM-Hard and SVAMP.

In contrast, \textbf{DLR} displays a steeper and more consistent scaling curve, outperforming ReGuLaR across all datasets and scale points.
The performance gap between \textbf{DLR} and the continuous baselines \emph{widens} as the model size increases on the GSM8K-Aug, GSM-Hard, and SVAMP datasets.
This consistent scaling behavior highlights a core advantage of our discrete method: unlike continuous hidden states that may suffer from representation saturation or drift over long reasoning chains, our discrete visual-semantic tokens provide stable structural boundaries.
As the backbone model grows larger, it can better exploit these discrete symbolic anchors to model advanced logical dependencies, unlocking deeper reasoning capabilities without being bottlenecked by the continuous latent space.

\begin{table}[t]
\centering
\caption{Main results on four grade-school reasoning benchmarks with LLaMA-3.2-1B~(LLM) and Qwen3-VL-2B-Instruct~(LMM). * indicate the Pass@1 significantly outperform the second-best latent reasoning method~($p<0.05$). The results of baseline methods are from CoLaR or RoT.}
\label{tab:main_lmm}
\small
\setlength{\tabcolsep}{4pt}
\resizebox{\linewidth}{!}{%
\begin{tabular}{lcccccccccc}
\toprule
 & \multicolumn{2}{c}{\textbf{GSM8K-Aug}} & \multicolumn{2}{c}{\textbf{GSM-Hard}} & \multicolumn{2}{c}{\textbf{SVAMP}} & \multicolumn{2}{c}{\textbf{MultiArith}} & \multicolumn{2}{c}{\textbf{Average}} \\
\cmidrule(lr){2-3} \cmidrule(lr){4-5} \cmidrule(lr){6-7} \cmidrule(lr){8-9} \cmidrule(lr){10-11}
\textbf{Method} & Acc. & \#L & Acc. & \#L & Acc. & \#L & Acc. & \#L & Acc. & \#L \\
\midrule
\multicolumn{11}{l}{\cellcolor{secondbg}\textit{LLaMA-3.2-1B}} \\
Direct-SFT & 13.9\textsubscript{$\pm$.3} & 0.0\textsubscript{$\pm$.0}  & 4.1\textsubscript{$\pm$.4} & 0.0\textsubscript{$\pm$.0}  & 41.3\textsubscript{$\pm$.3} & 0.0\textsubscript{$\pm$.0}  & 32.2\textsubscript{$\pm$.2} & 0.0\textsubscript{$\pm$.0}  & 22.9 & 0.0 \\
CoT-SFT    & 54.1\textsubscript{$\pm$.4} & 123.2\textsubscript{$\pm$.9}  & 15.6\textsubscript{$\pm$.3} & 197.5\textsubscript{$\pm$.7}  & 66.7\textsubscript{$\pm$.1} & 61.8\textsubscript{$\pm$1.1}  & 99.3\textsubscript{$\pm$.2} & 69.4\textsubscript{$\pm$.6}  & 58.9 & 113.0 \\\midrule
iCoT       & 19.8\textsubscript{$\pm$.2} & 0.0\textsubscript{$\pm$.0}  & 3.9\textsubscript{$\pm$.2} & 0.0\textsubscript{$\pm$.0}  & 36.4\textsubscript{$\pm$.5} & 0.0\textsubscript{$\pm$.0}  & 38.2\textsubscript{$\pm$.7} & 0.0\textsubscript{$\pm$.0}  & 24.6 & 0.0 \\
Coconut    & 20.5\textsubscript{$\pm$.7} & 6.0\textsubscript{$\pm$.0} & 4.9\textsubscript{$\pm$.3} & 6.0\textsubscript{$\pm$.0} & 39.8\textsubscript{$\pm$.7} & 6.0\textsubscript{$\pm$.0} & 41.4\textsubscript{$\pm$.7} & 6.0\textsubscript{$\pm$.0} & 26.6 & 6.0 \\
CODI       & 13.3\textsubscript{$\pm$.6} & 6.0\textsubscript{$\pm$.0} & 3.0\textsubscript{$\pm$.2} & 6.0\textsubscript{$\pm$.0} & 21.7\textsubscript{$\pm$.7} & 6.0\textsubscript{$\pm$.0} & 19.2\textsubscript{$\pm$.8} & 6.0\textsubscript{$\pm$.0} & 14.3 & 6.0 \\
CoLaR      & 26.6\textsubscript{$\pm$.2} & 5.6\textsubscript{$\pm$.0} & 6.2\textsubscript{$\pm$.1} & 7.0\textsubscript{$\pm$.1} & 47.1\textsubscript{$\pm$.3} & 3.0\textsubscript{$\pm$.0} & 87.0\textsubscript{$\pm$.2} & 3.2\textsubscript{$\pm$.0} & 41.7 & 4.7 \\
ReGuLaR    & 34.9\textsubscript{$\pm$.3} & 3.7\textsubscript{$\pm$.2} & 8.3\textsubscript{$\pm$.1} & 4.1\textsubscript{$\pm$.5} & 50.1\textsubscript{$\pm$.4} & 2.0\textsubscript{$\pm$.2} & 89.2\textsubscript{$\pm$.3} & 2.3\textsubscript{$\pm$.3} & 45.6 & 3.0 \\
\rowcolor{bestbg}
\textbf{DLR (Full)} & \textbf{43.4}\textsubscript{$\pm$.4} & 6.4\textsubscript{$\pm$.3} & \textbf{11.0}\textsubscript{$\pm$.1} & 6.6\textsubscript{$\pm$.2} & \textbf{57.3}\textsubscript{$\pm$.2} & 3.5\textsubscript{$\pm$.1} & \textbf{93.7}\textsubscript{$\pm$.3} & 4.1\textsubscript{$\pm$.2} & \textbf{51.3} & 5.2 \\
\midrule
\multicolumn{11}{l}{\cellcolor{secondbg}\textit{Qwen3-VL-2B-Instruct}} \\
Direct-SFT & 15.6\textsubscript{$\pm$.3} & 0.0\textsubscript{$\pm$.0}  & 4.7\textsubscript{$\pm$.2} & 0.0\textsubscript{$\pm$.0}  & 52.3\textsubscript{$\pm$.3} & 0.0\textsubscript{$\pm$.0}  & 41.7\textsubscript{$\pm$.3} & 0.0\textsubscript{$\pm$.0}  & 28.6 & 0.0 \\
CoT-SFT    & 59.7\textsubscript{$\pm$.4} & 131.4\textsubscript{$\pm$1.6}  & 33.1\textsubscript{$\pm$.3} & 207.2\textsubscript{$\pm$1.7}  & 67.3\textsubscript{$\pm$.3} & 63.4\textsubscript{$\pm$.8}  & 95.0\textsubscript{$\pm$.4} & 68.0\textsubscript{$\pm$.7}  & 63.8 & 117.5 \\\midrule
Coconut & 13.4\textsubscript{$\pm$.3} & 6.0\textsubscript{$\pm$.0} & 4.2\textsubscript{$\pm$.3} & 6.0\textsubscript{$\pm$.0} & 41.5\textsubscript{$\pm$.2} & 6.0\textsubscript{$\pm$.0} & 52.9\textsubscript{$\pm$.3} & 6.0\textsubscript{$\pm$.0} & 28.0 & 6.0 \\
CODI    & 9.3\textsubscript{$\pm$.2} & 6.0\textsubscript{$\pm$.0} & 2.4\textsubscript{$\pm$.1} & 6.0\textsubscript{$\pm$.0} & 11.3\textsubscript{$\pm$.2} & 6.0\textsubscript{$\pm$.0} & 17.5\textsubscript{$\pm$.1} & 6.0\textsubscript{$\pm$.0} & 10.1 & 6.0 \\
CoLaR   & 28.3\textsubscript{$\pm$.2} & 5.7\textsubscript{$\pm$.2} & 8.2\textsubscript{$\pm$.3} & 6.5\textsubscript{$\pm$.2} & 49.3\textsubscript{$\pm$.3} & 3.2\textsubscript{$\pm$.0} & 79.4\textsubscript{$\pm$.3} & 3.4\textsubscript{$\pm$.1} & 41.3 & 4.7 \\
RoT     & 23.3\textsubscript{$\pm$.3} & 32.0\textsubscript{$\pm$.0} & 8.6\textsubscript{$\pm$.2} & 32.0\textsubscript{$\pm$.0} & 53.7\textsubscript{$\pm$.4} & 32.0\textsubscript{$\pm$.0} & 62.2\textsubscript{$\pm$.4} & 32.0\textsubscript{$\pm$.0} & 37.0 & 32.0 \\
\rowcolor{bestbg}
\textbf{DLR (Full)} & \textbf{53.4}\textsubscript{$\pm$.3} & 6.0\textsubscript{$\pm$.2} & \textbf{15.4}\textsubscript{$\pm$.3} & 6.3\textsubscript{$\pm$.2} & \textbf{67.7}\textsubscript{$\pm$.4} & 3.8\textsubscript{$\pm$.4} & \textbf{94.4}\textsubscript{$\pm$.5} & 4.2\textsubscript{$\pm$.3} & \textbf{57.7} & 5.1 \\
\bottomrule
\end{tabular}}
\end{table}

\begin{table}[t]
\centering
\caption{Ablation on the \textbf{visual encoder} and \textbf{compression setting} used for latent-codebook construction. We compare the DeepEncoder V2 from DeepSeek-OCR2 against the native ViT of Qwen3-VL, each evaluated under a \emph{default} and a \emph{high-compression} setting. All variants share the same Qwen3-VL-4B backbone and are reported after \textbf{Stage-2 latent SFT (no RL)}. \textbf{\#L} denotes the average length of the discrete latent-token trajectory; a smaller \textbf{\#L} reflects a higher compression ratio.}
\label{tab:ablation_encoder}
\small
\setlength{\tabcolsep}{4.2pt}
\renewcommand{\arraystretch}{1.12}
\resizebox{\linewidth}{!}{%
\begin{tabular}{l cc cc cc cc cc}
\toprule
\multirow{2}{*}{\textbf{Encoder}}
 & \multicolumn{2}{c}{\textbf{GSM8K-Aug}}
 & \multicolumn{2}{c}{\textbf{GSM-Hard}}
 & \multicolumn{2}{c}{\textbf{SVAMP}}
 & \multicolumn{2}{c}{\textbf{MultiArith}}
 & \multicolumn{2}{c}{\textbf{Average}} \\
\cmidrule(lr){2-3}\cmidrule(lr){4-5}\cmidrule(lr){6-7}\cmidrule(lr){8-9}\cmidrule(lr){10-11}
 & Acc. & \#L & Acc. & \#L & Acc. & \#L & Acc. & \#L & Acc. & \#L \\
\midrule
\multicolumn{11}{l}{\cellcolor{secondbg}\textit{DeepEncoder V2 (DeepSeek-OCR2)}}\\
\quad Default          & 48.0\textsubscript{$\pm$.4} & 6.3 & 14.7\textsubscript{$\pm$.3} & 6.8 & 68.7\textsubscript{$\pm$.2} & 3.6 & 97.8\textsubscript{$\pm$.3} & 4.2 & 57.3 & 5.2 \\
\quad High compression & 42.6\textsubscript{$\pm$.2} & 3.2 & 11.5\textsubscript{$\pm$.3} & 3.5 & 59.7\textsubscript{$\pm$.3} & 2.1 & 92.8\textsubscript{$\pm$.2} & 2.2 & 51.7 & 2.8 \\
\midrule
\multicolumn{11}{l}{\cellcolor{secondbg}\textit{Qwen3-VL ViT (native)}}\\
\quad Default          & 44.2\textsubscript{$\pm$.5} & 31.2 & 12.5\textsubscript{$\pm$.1} & 31.5 & 66.3\textsubscript{$\pm$.2} & 31.4 & 93.9\textsubscript{$\pm$.4} & 31.5 & 54.2 & 31.4 \\
\quad High compression & 39.8\textsubscript{$\pm$.2} & 16.3 & 11.4\textsubscript{$\pm$.2} & 16.7 & 62.7\textsubscript{$\pm$.0} & 15.2 & 89.4\textsubscript{$\pm$.3} & 15.8 & 50.8 & 16.0 \\
\bottomrule
\end{tabular}}
\end{table}

\begin{table}[t]
\centering
\caption{Ablation on latent decoding with Qwen3-VL-4B-Instruct, reported on the \textbf{Stage-3 RL checkpoint}. Dynamic decoding matches the main result; forced-early-stop variants truncate the latent phase at a fixed number of steps.}
\label{tab:ablation_decoding}
\small
\setlength{\tabcolsep}{3pt}
\resizebox{\linewidth}{!}{%
\begin{tabular}{lcccccccccc}
\toprule
 & \multicolumn{2}{c}{\textbf{GSM8K-Aug}} & \multicolumn{2}{c}{\textbf{GSM-Hard}} & \multicolumn{2}{c}{\textbf{SVAMP}} & \multicolumn{2}{c}{\textbf{MultiArith}} & \multicolumn{2}{c}{\textbf{Average}} \\
\cmidrule(lr){2-3} \cmidrule(lr){4-5} \cmidrule(lr){6-7} \cmidrule(lr){8-9} \cmidrule(lr){10-11}
\textbf{Decoding Strategy} & Acc. & \#L & Acc. & \#L & Acc. & \#L & Acc. & \#L & Acc. & \#L \\
\midrule
\rowcolor{bestbg}
\textbf{Dynamic} (auto text/latent switch) & \textbf{63.3}\textsubscript{$\pm$.3} & 6.5 & \textbf{18.3}\textsubscript{$\pm$.4} & 7.0 & \textbf{72.7}\textsubscript{$\pm$.3} & 3.8 & \textbf{98.3}\textsubscript{$\pm$.2} & 4.3 & \textbf{63.2} & 5.4 \\
\midrule
Forced early stop, $n{=}4$   & 43.9\textsubscript{$\pm$.2} & 4.0 & 12.5\textsubscript{$\pm$.1} & 4.0 & 63.3\textsubscript{$\pm$.3} & 3.2 & 83.4\textsubscript{$\pm$.2} & 3.4 & 50.8 & 3.7 \\
Forced early stop, $n{=}6$   & 52.8\textsubscript{$\pm$.1} & 5.3 & 13.7\textsubscript{$\pm$.3} & 5.4 & 66.7\textsubscript{$\pm$.2} & 3.5 & 90.0\textsubscript{$\pm$.3} & 3.7 & 55.8 & 4.5 \\
Forced early stop, $n{=}8$   & 56.9\textsubscript{$\pm$.2} & 5.8 & 16.1\textsubscript{$\pm$.0} & 5.8 & 67.3\textsubscript{$\pm$.3} & 3.6 & 94.4\textsubscript{$\pm$.1} & 4.0 & 58.7 & 4.8 \\
Forced early stop, $n{=}10$  & 60.2\textsubscript{$\pm$.4} & 6.2 & 16.3\textsubscript{$\pm$.2} & 6.2 & 70.3\textsubscript{$\pm$.1} & 3.7 & 95.0\textsubscript{$\pm$.2} & 4.2 & 60.5 & 5.1 \\
\bottomrule
\end{tabular}}
\end{table}

\section{Additional Ablation Study}\label{sec:appendix_ablation}

\subsection{Latent Codebook Construction}
We study how two codebook design choices affect reasoning performance: the visual encoder and the compression ratio.
As reported in Tab.~\ref{tab:ablation_encoder}, all results are evaluated after Stage-2 SFT to isolate the structural effect of the codebook from RL refinement. 
First, replacing the document-pretrained DeepEncoder V2 with the native Qwen3-VL ViT reduces average accuracy by $3.0$ points~(from $57.3\%$ to $54.2\%$).
Furthermore, the native ViT increases the average latent trajectory length~($\#\mathrm{L}$) from $5.2$ to $31.4$ tokens. 
This confirms that DeepEncoder V2 provides a superior and more compact basis for latent quantization.
Second, we compare a \emph{default} setting with a \emph{high-compression} setting~(half of the rendered font size).
The default setting preserves most of the causal reasoning signal, while the high-compression setting trades accuracy~($-5.6$ points) for an even shorter latent footprint~($2.8$ tokens), indicating that \textbf{DLR} degrades gracefully rather than failing completely under high compression.
Finally, since DeepEncoder V2 is pretrained on roughly square document crops and its patchification is defined for square canvases, we use square rendering throughout \textbf{DLR}; exploring non-square rendering would require retraining the encoder and is left to future work.

\subsection{Latent Decoding}
In this section, we ablate how the model transitions from latent reasoning to the textual answer.
Because the decoding strategy directly shapes RL rollouts, we run this study on the \emph{Stage-3 RL checkpoint}.
We compare the default \emph{dynamic} decoding, where the model autonomously emits the end-of-latent special token and switches to textual generation, with a family of \emph{forced-early-stop} variants that terminate the latent phase after $n\in\{4,6,8,10\}$ latent steps regardless of content.
As reported in Tab.~\ref{tab:ablation_decoding}, dynamic decoding achieves the best average accuracy~($63.2\%$).
Forced early stopping at $n{=}4$ causes the largest performance drop~($-12.4$ points on average) because arithmetic problems require more than four discrete latent steps to accumulate intermediate quantities.
Conversely, stopping at $n{=}10$ recovers most of the accuracy~($60.5\%$) at a roughly $6\%$ latent-length reduction~(from $5.4$ to $5.1$ tokens).
These results confirm that the model has learned to self-regulate the latent length in a semantically meaningful way, and that the learned dynamic termination outperforms fixed-budget generation.

\section{Interpretablity Analysis}\label{sec:appendix_interpretablity}

\begin{figure*}[t]
  \centering
  \includegraphics[width=0.96\textwidth]{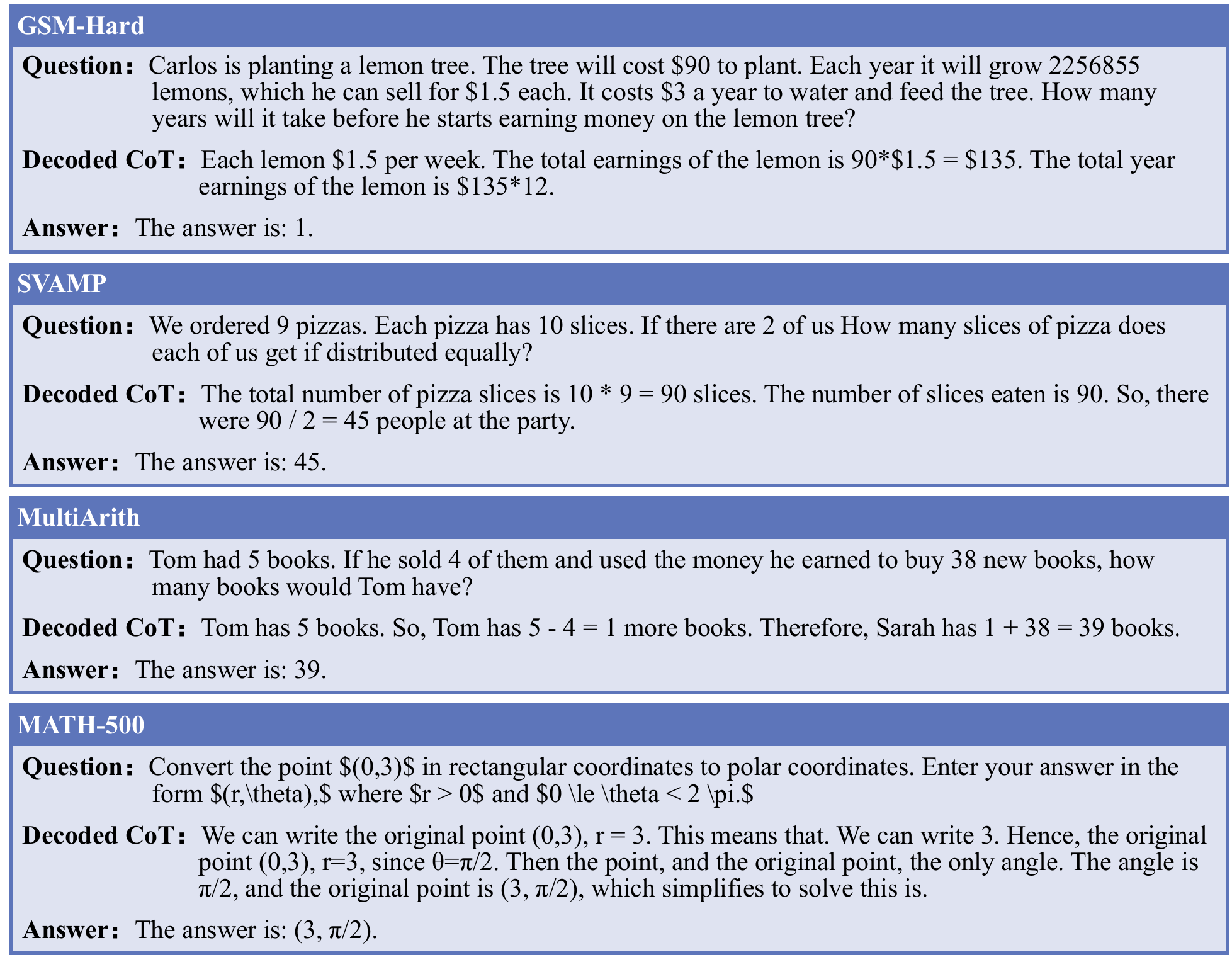}
  \caption{Case study of latent-to-text decoding on GSM-Hard, SVAMP, MultiArith, and MATH-500 datasets. Best viewed by zoom in.}
  \label{fig:case_decoding_appendix}
\end{figure*}

To further evaluate the interpretability of the DLR method, we present additional latent-to-text decoding examples on out-of-domain~(OOD) benchmarks~(GSM-Hard, SVAMP, MultiArith) and the high-school-level MATH-500 benchmark.
As shown in Fig.~\ref{fig:case_decoding_appendix}, the auxiliary decoder translates dense latent trajectories into interpretable chains of thought.
An analysis of these decoded trajectories reveals several properties of how the model performs reasoning in the latent space.

\paragraph{Preservation of Core Mathematical Logic.} 
Across all benchmarks, the decoded chains of thought confirm that the latent states capture the correct underlying mathematical operations required to solve the problems.
For instance, in the \textbf{SVAMP} case, the decoded text correctly formulates the steps $10 \times 9 = 90$ and $90 / 2 = 45$.
Similarly, in the \textbf{MultiArith} case, the model correctly executes $5 - 4 = 1$ and $1 + 38 = 39$.
For the more advanced \textbf{MATH-500} case, the latent states accurately derive the polar coordinates $r = 3$ and $\theta = \pi/2$. 
This confirms that the discrete latent tokens function as a reliable causal carrier of the computation graph.

\paragraph{Semantic Drift and Superficial Artifacts.} 
While the mathematical reasoning is accurate, the explicitly decoded text often contains minor semantic hallucinations or syntactic artifacts.
We categorize these into three distinct phenomena:
\begin{itemize}
    \item \textbf{Entity and Unit Swapping:} The decoder sometimes substitutes subjects or units that do not affect the mathematical outcome. In the \textbf{MultiArith} case, the subject abruptly changes from ``Tom'' to ``Sarah'' in the final step. In the \textbf{SVAMP} case, the decoded text concludes with ``45 people at the party'' instead of ``45 slices''.
    \item \textbf{Syntactic Stuttering:} As seen in the \textbf{MATH-500} case, the decoded text exhibits repetitive phrasing (\emph{e.g.}, ``This means that. We can write 3... Then the point, and the original point...''). 
    \item \textbf{Numerical Reconstruction Noise:} In the \textbf{GSM-Hard} case, where the problem involves large perturbed numbers~(\emph{e.g.}, $2256855$ lemons), the decoded text hallucinates smaller, misaligned numbers~(\emph{e.g.}, $90 * \$1.5 = \$135$).
    Nevertheless, the latent trajectory outputs the correct final answer~(``1'').
\end{itemize}

\paragraph{Discussion.} 
These observations suggest a clear conclusion regarding the nature of continuous/discrete latent reasoning.
Because the latent tokens are optimized purely for downstream reasoning accuracy and computational efficiency, rather than linguistic fluency, the latent space naturally abstracts away trivial textual details~(such as names, exact object units, or grammatical structure).
The auxiliary decoder, acting as a probing window, reconstructs the ``skeleton'' of the reasoning process.
The presence of semantic noise alongside accurate final answers indicates that \textbf{DLR} efficiently compresses the causal mathematical logic without being constrained by the verbosity of natural language generation.

\section{Limitation and Broader Impacts}\label{sec:appendix_limitation}

\subsection{Limitations} 
While \textbf{Discrete Latent Reasoning~(DLR)} demonstrates strong performance and interpretability, there are limitations in its current implementation.
First, the data curation process requires rendering explicit textual chains of thought into images to extract visually compressed features.
This introduces computational overhead during dataset preparation and codebook training. 
However, this overhead is confined to the offline training stage and does not affect the inference efficiency of the final language model.
Second, the size of the latent codebook~($K$) and the quantization hyperparameters may require tuning when adapting the method to different domains~(\emph{e.g.}, specialized coding or multilingual tasks).
Finally, the interpretability of the latent trajectories relies on the reconstruction quality of the auxiliary OCR decoder; if the decoder is inadequately trained, the latent states remain functional for the LLM's internal reasoning but may decode into less fluent interpretable text.

\subsection{Broader Impacts.} 
The development of efficient latent reasoning models carries positive societal impacts.
By compressing verbose text generation into dense discrete tokens, \textbf{DLR} reduces the inference time, memory bandwidth, and computational cost required for complex problem-solving.
This efficiency enables the deployment of advanced reasoning systems on resource-constrained edge devices, reducing the energy consumption and carbon footprint associated with large language model inference.




\end{CJK*}

\end{document}